\documentclass{article}

\usepackage[preprint]{neurips_2022}

\usepackage[utf8]{inputenc} % allow utf-8 input
\usepackage[T1]{fontenc}    % use 8-bit T1 fonts
\usepackage{hyperref}       % hyperlinks
\usepackage{url}            % simple URL typesetting
\usepackage{booktabs}       % professional-quality tables
\usepackage{amsfonts}       % blackboard math symbols
\usepackage{amsmath}
\usepackage{nicefrac}       % compact symbols for 1/2, etc.
\usepackage{microtype}      % microtypography
\usepackage{xcolor}         % colors
\usepackage{todonotes}
\usepackage{subcaption}
\usepackage{multirow}
\usepackage{bbm}

\usepackage{amsmath}
\usepackage{graphicx}
\usepackage{subcaption}
\usepackage{tikz}
\usepackage{bm}
\usepackage{booktabs}
\usepackage{siunitx}
\usepackage{wrapfig}
\usepackage{enumitem}
\setlist[itemize]{align=parleft,left=0pt..1em}
\newcommand{\rpm}{\raisebox{.2ex}{$\scriptstyle\pm$}}
\setcitestyle{numbers,square}

\usepackage{amsthm}

\definecolor{dtblue}{rgb}{0.8, 0.8, 1}
\definecolor{dtgreen}{rgb}{0.8, 1, 0.8}
\definecolor{dtyellow}{rgb}{1, 1, 0.8}

\title{DT+GNN: A Fully Explainable Graph Neural Network using Decision Trees}

\author{%
Peter M\"uller\\
ETH Zurich, Switzerland\\
\texttt{pemuelle@ethz.ch}\\
\And
Lukas Faber\\
ETH Zurich, Switzerland\\
\texttt{lfaber@ethz.ch}\\
\AND
Karolis Martinkus\\
ETH Zurich, Switzerland\\
\texttt{martinkus@ethz.ch}\\
\And
Roger Wattenhofer\\
ETH Zurich, Switzerland\\
\texttt{wattenhofer@ethz.ch}\\
}

\begin{document}

\maketitle

\begin{abstract}
We propose the fully explainable Decision Tree Graph Neural Network (DT+GNN) architecture. In contrast to existing black-box GNNs and post-hoc explanation methods, the reasoning of DT+GNN can be inspected at every step. To achieve this, we first construct a differentiable GNN layer, which uses a categorical state space for nodes and messages. This allows us to convert the trained MLPs in the GNN into decision trees. These trees are pruned using our newly proposed method to ensure they are small and easy to interpret. We can also use the decision trees to compute traditional explanations. We demonstrate on both real-world datasets and synthetic GNN explainability benchmarks that this architecture works as well as traditional GNNs. Furthermore, we leverage the explainability of DT+GNNs to find interesting insights into many of these datasets, with some surprising results. We also provide an interactive web tool to inspect DT+GNN's decision making.
\end{abstract}

\section{Introduction}
Graph Neural Networks (GNNs) have been successful in applying machine learning techniques to many graph-based domains~\citep{bian2020rumor, gilmer2017neural, yang2020financial, jumper2021highly}. However, currently, GNNs are black-box models,
\begin{wrapfigure}{r}{0.37\textwidth}
\centering
    \includegraphics[width=0.35\textwidth]{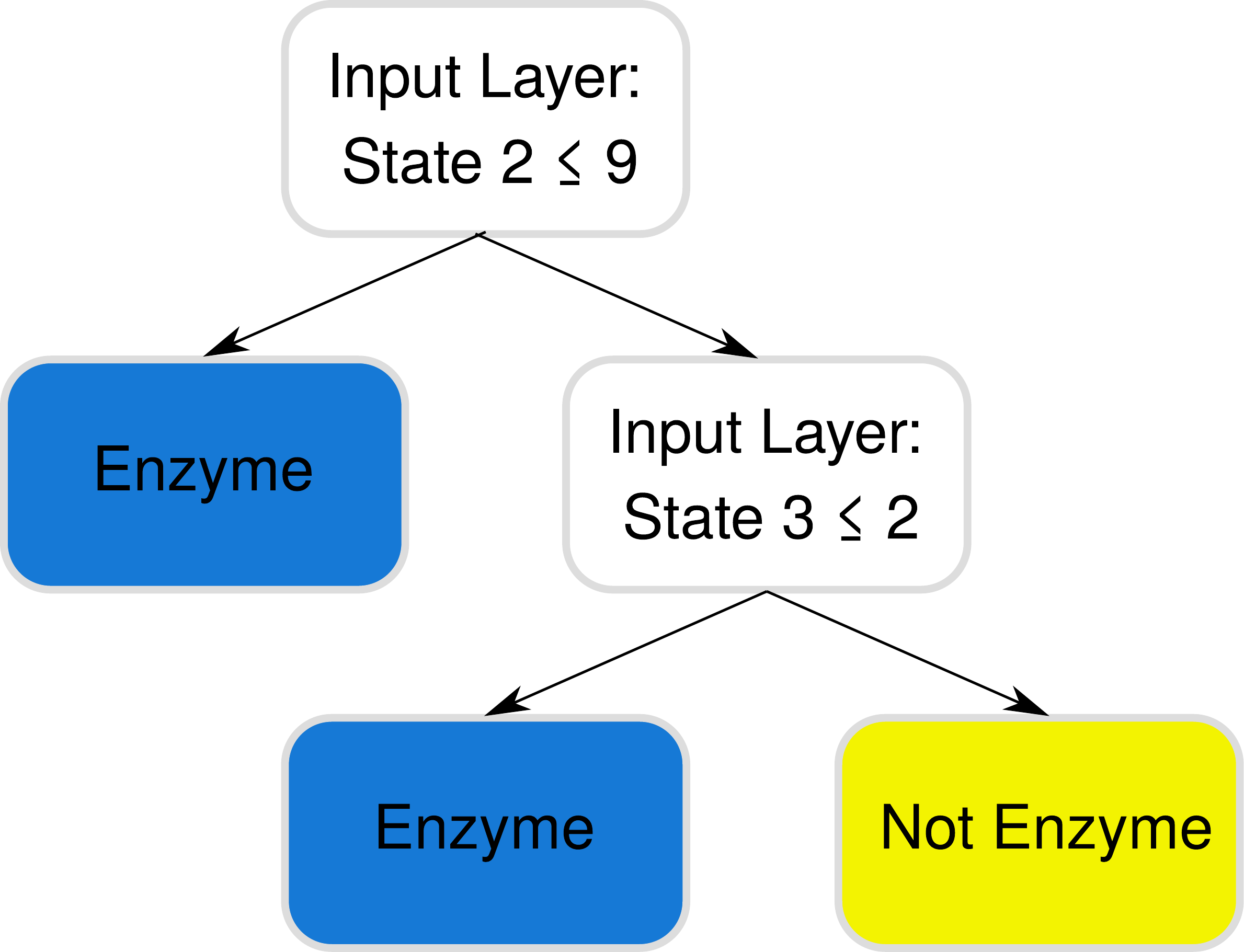}
    \caption{Final layer decision rule of DT+GNN for the PROTEINS dataset. This layer counts how many nodes representing certain chemical structures (states 2 and 3) the protein has. If there are not too many, the protein is an enzyme.}
    \label{fig:interpretation}
    \vspace{-2em} % otherwise whitespace overflows to the next page - the beauty of wrapfig
\end{wrapfigure}
and their lack of human interpretability limits their use in many potential application areas.
This motivated the adoption of existing deep learning explanation methods~\citep{baldassarre2019explainability, huang2020graphlime, pope2019explainability} to GNNs, and the creation of new GNN-specific methods~\citep{ying2019gnnexplainer, vu2020pgmexplainer}. These methods allow us to understand \emph{what} parts of the input were important for making a prediction. However, these methods do not explain \emph{how} a GNN uses its inputs.

In this paper, we present a new fully explainable GNN architecture called Decision Tree GNN (DT+GNN). Fully explainable means that we can also inspect the decision making process and see \textit{how} the GNN uses its inputs. For example, consider the output layer of DT+GNN for the PROTEINS dataset~\cite{Borgwardt2005ProteinKernels} in Figure~\ref{fig:interpretation}. We see that DT+GNN simply counts the occurrences of certain chemical structures in the input graph to make its prediction. DT+GNN does not care at all how the amino acids are connected. In fact, this confirms a known observation that simple counting can outperform traditional GNNs on this dataset~\citep{errica2020fair}. DT+GNN provides insights not only for PROTEINS but also for many other datasets.
We discuss these for the BA-2Motifs~\citep{luo2020pgexplainer}, Tree-Cycle~\citep{ying2019gnnexplainer}, and MUTAG\citep{debnath1991structure} datasets in Section~\ref{sec:example_interpretations}, and others in Appendix~\ref{sec:more_example_interpretations}. We summarize our contributions as follows:
\begin{itemize}
    \item We propose a new differentiable Diff-DT+GNN layer. While traditional GNNs are based on a distributed computing model known as synchronous message passing  \citep{loukas2019gnnlocal}, our new layer is based on a simplified distributed computing model known as the stone age model \citep{emek2013stoneage}. In this model, nodes use a small categorical space for states and messages. We argue that the stone age model is more suitable for interpretation while retaining a high theoretical expressiveness.
    \item We make our model fully interpretable by making use of decision trees. We first train Diff-DT+GNN using gradient descent. Internally, Diff-DT+GNN uses Multi-Layer Perceptrons (MLPs). Then, we replace each MLP with a decision tree while keeping the original GNN message passing structure. This gives us a model consisting of a series of decision tree layers, which makes it fully interpretable. We name our model Decision Tree GNN (DT+GNN).
    \item We propose a way to collectively prune the decision trees in DT+GNN without compromising accuracy. This leads to smaller trees, which further increases explainability.
    \item We further provide a way to extract traditional GNN explanations from DT+GNN. These explanations form a heat map that highlights the important nodes in the graph.
    % which helps human inspection and interpretation. 
    % We also propose derive a scheme for computing explanations with DT+GNN to guide human users towards the important areas.
    \item We test our proposed architecture on established GNN explanation benchmarks and real-world graph datasets. We show our models are competitive in classification accuracy with traditional GNNs. We further validate that the proposed pruning methods considerably reduce tree sizes. We also demonstrate that DT+GNN can be used to discover problems in existing explanation benchmarks and to find interesting insights into real-world datasets.
    \item Finally, we provide a user interface for DT+GNN.\footnote{\url{https://interpretable-gnn.netlify.app/}} This tool allows for the interactive exploration of the DT+GNN decision process on the datasets examined in this paper. We provide a manual for the interface in Appendix~\ref{sec:manual}.
\end{itemize}

\section{Related Work}
\subsection{Explanation methods for GNNs}
\label{sec:explanation_methods}

In recent years, several methods for providing GNN explanations were proposed. These methods highlight which parts of the input are important in a GNN decision. Generally, they explain a prediction either by assigning importance scores to nodes and edges or by finding similar predictions to help humans recognize patterns. The existing explanation methods can be roughly grouped into the following five groups:

\textbf{Gradient based.} \citet{baldassarre2019explainability} and \citet{pope2019explainability} show that it is possible to adopt gradient-based methods that we know from computer vision, for example Grad-CAM\citep{selvaraju2017grad}, to graphs. Gradients can be computed on node features and edges~\citep{schlichtkrull2021interpreting}.\\
\textbf{Mutual-Information based.} \citet{ying2019gnnexplainer} also measure the importance of edges and node features. Edges are masked with continuous values. Instead of gradients, the authors use mutual information between inputs and the prediction to quantify the impact. \citet{luo2020pgexplainer} follow a similar idea but emphasize finding important edges and finding explanations for many predictions at the same time.\\
\textbf{Subgraph based.} \citet{yuan2021subgraphx} search the space of all subgraphs as possible explanations. To score a subgraph, they use Shapley values~\citep{shapley1953value} and monte carlo tree search for guiding the search. \citet{duval2021graphsvx} build subgraphs by masking nodes and edges in the graph. They run their subgraph through the trained GNN and try to explain the differences to the entire graph with simple interpretable models and Shapley values. \citet{zhang2021protgnn} infer subgraphs called prototypes that each map to one particular class. Graphs are classified through their similarity to the prototypes.\\
\textbf{Example based.} \citet{huang2020graphlime} proposes a graph version of the LIME~\citep{ribeiro2016should} algorithm. A prediction is explained through a linear decision boundary built by close-by examples. \citet{vu2020pgmexplainer} aim to capture the dependencies in node predictions and express them in probabilistic graphical models. \citet{faber2020contrastive} explain a node by giving examples of similar nodes with the same and different labels. \citet{dai2021selfexplainable} create a $k$-nearest neighbor model and use GNNs to create the feature space to compute similarities in.\\
\textbf{Simple GNNs.} Another interesting line of research are simplified GNN architectures \cite{cai2018classification, chen2019powerful, Huang2020Combining}. The main goal of this research is to show that simple architectures can perform competitively with traditional, complex ones. As a side effect, the simplicity of these architectures also makes them slightly more understandable.

Complimentary to the development of explanations methods is the research on how we can best evaluate them. \citet{sanchez2020evaluating} and \citet{yuan2020explainabilitySurvey} discuss desirable properties a good explanation method should have. For example, an explanation method should be faithful to the model, which means that an explanation method should reflect the model's performance and behavior. \citet{agarwal2022probing} provide a theoretical framework to define how strong explanation methods adhere to these properties. They also derive bounds for several explanation methods. \citet{faber2021comparing} and \citet{himmelhuber2021demystifying} discuss deficiencies in the existing benchmarks used for empirical evaluation. %of explanation methods.

Note that all these methods do not provide \textit{full} explainability. They can show us the important nodes and edges a GNN uses to predict. But (in contrast to DT+GNN) they cannot give insights into the GNN's decision making process itself.

%Related to this, another interesting line of research are simplified GNN architectures \cite{chen2019powerful, cai2018classification, Huang2020Combining}. The main goal of this research is to show that simple architectures can perform competitively to the traditional, complex ones. Due to their simplicity these architectures are also slightly more interpretable. Unfortunately, none of the previous work focused on building an architecture that is truly interpretable. With our DT+GNN we show that it is possible to build a model that is fully interpretable and essentially keeps the expressiveness of GNNs.

\subsection{Decision trees for neural networks}
Decision trees are powerful machine learning methods, which in some tasks rival the performance of neural networks \citep{gorishniy2021revisiting}. The cherry on top is their inherent interpretability. Therefore, researchers have tried to distill neural networks into trees to make them explainable~\citep{boz2002extracting, dancey2004decision, krishnan1999extracting}. More recently, \citet{schaaf2019enhancing} have shown that encouraging sparsity and orthogonality in neural network weight matrices allows for model distillation into smaller trees with higher final accuracy. \citet{wu2017beyond} follow a similar idea for time series data: they regularize the training process for recurrent neural networks to penalize weights that cannot be modeled easily by complex decision trees.
\citet{yang2018deepneural} aim to directly learn neural trees. Their neural layers learn how to split the data and put it into bins. Stacking these layers creates trees. \citet{kontschieder2015deep} learn neural decision forests by making the routing in nodes probabilistic and learning these probabilities and the leaf predictions.

Our DT+GNN follows the same underlying idea: we want to structure a GNN in a way that allows for model distillation into decision trees to leverage their interpretability. However, the graph setting brings additional challenges. We not only have feature matrices for each node, but we also need to allow the trees to reason about the state of neighboring nodes one or more hops away.
 
\section{The DT+GNN Model}

\subsection{The Diff-DT+GNN layer}\label{sec:stone_age}
\citet{loukas2019gnnlocal} show that GNNs operate very similar to synchronous message passing algorithms from distributed computing. Often, these algorithms have a limit on the message size of $b=O(\log n)$ bits (where $n$ is the number of nodes) but can perform arbitrary local computation~\citep{peleg2000distributed}. In contrast to this, the stone age distributed computing model \citep{emek2013stoneage} assumes that each node uses a finite state machine to update its state and send its updated state as a message to all neighbors. The receiving node can count the number of neighbors in each state. A stone age node cannot even count arbitrarily, it can only count up to a predefined number, in the spirit of ``one, two, many''. Neighborhood counts above a threshold are indistinguishable from each other. Interestingly enough, such a simplified model can still solve many distributed computing problems~\citep{emek2013stoneage}.

\begin{figure}
    \centering
    \includegraphics[width=0.9\textwidth]{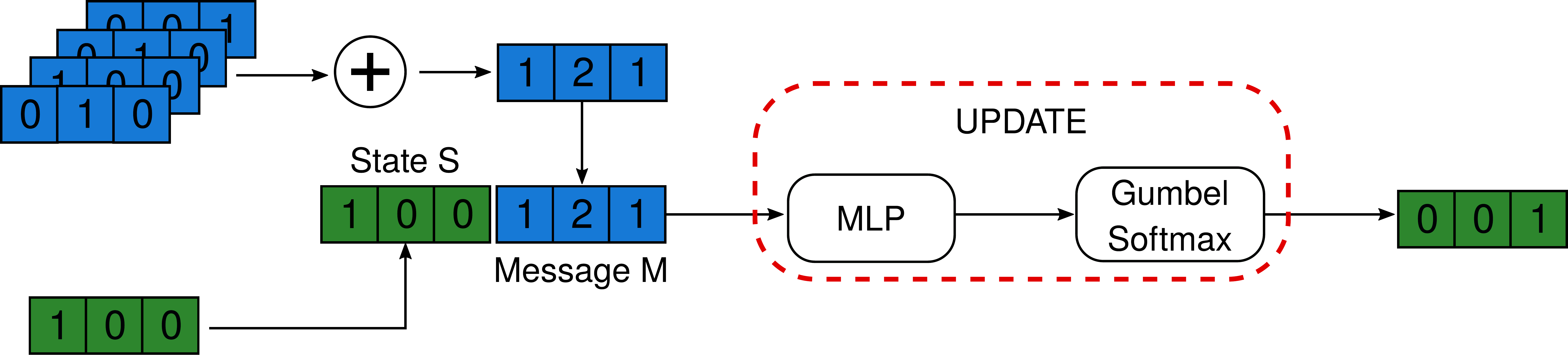}
    \caption{A Diff-DT+GNN layer. It follows normal message passing layers to a large extend, except it computes a categorical state in the end, instead of a real-valued embedding vector.}
    \label{fig:stone_age_layer}
\end{figure}

Clearly, this simplified model would also be easier to interpret. We create the Diff-DT+GNN layer similar to this model (Figure~\ref{fig:stone_age_layer}) which is fully differentiable. The layer largely follows traditional message passing \citep{gilmer2017neural}. Nodes update their state through an \texttt{UPDATE} function that takes the aggregated messages $M$ from neighbors and the node's current state $S$ as input. To retain differentiability, this layer uses MLPs instead of finite state machines in the \texttt{UPDATE} step. We constrain the output space of the \texttt{UPDATE} step to be categorical, by using a  Gumbel-Softmax~\citep{jang2017gumbel, maddison2016concrete}. Similarly to models like GIN \citep{xu2018powerful}, every node sends its state directly to its neighbors. Together with \texttt{sum} aggregation, this allows each node to count neighbor states, as in the stone age model. However, we deviate from the stone age model by not restricting the neighborhood counting. We find this produces better results without sacrificing interpretability.

In principle, the categorical state space also does not reduce expressiveness. Using a GNN with messages that have $O(\log{n})$ bits is equivalent to using categorical messages with $O(n)$ bits. Practically, we are interested in small interpretable state spaces, and hence, we constrain the state space to small values. The reduced expressiveness does not impact the performance in practice (c.f., Table~\ref{tab:accuracies}).

We can now build Diff-DT+GNN from these layers. We add an encoder MLP at the beginning that projects the input features into a categorical space. We append a decoder MLP at the end that works on a concatenation of intermediate outputs. For node classification, the decoder concatenates node states; for graph classification, the decoder concatenates the pooled state counts.

\begin{figure}
    \centering
    \includegraphics[width=0.9\textwidth]{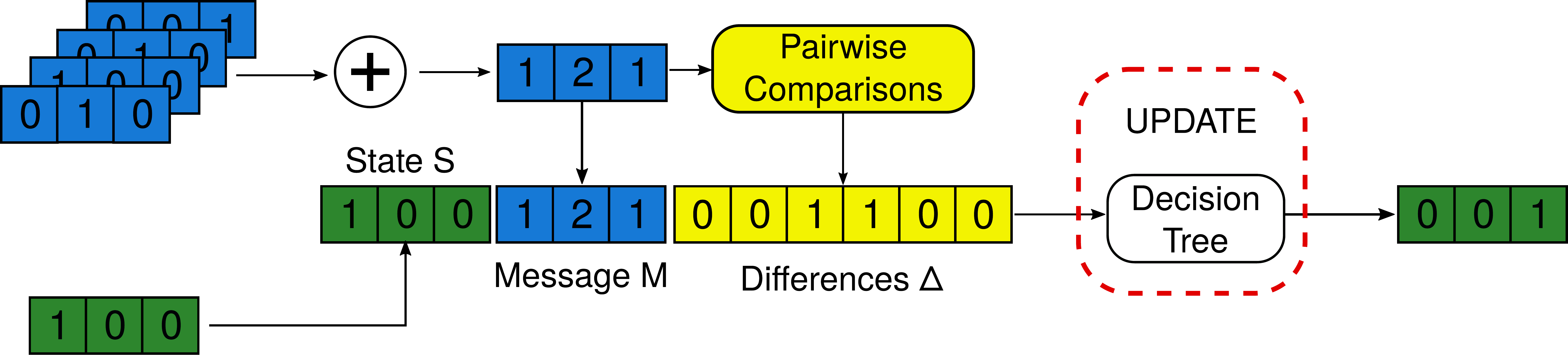}
    \caption{A DT+GNN layer. We replace the UPDATE of a Diff-DT+GNN layer with a decision tree. DT+GNN receive additional $\Delta$ features to compare two features.}
    \label{fig:dt_layer}
\end{figure}

\subsection{Distilling the DT+GNN}
Using an MLP in the \texttt{UPDATE} step harms explainability. To fix this, we distill all of the MLPs used in the GNN, including the encoder and decoder MLPs, into decision trees. To train the decision trees, we pass all of the training graphs through the Diff-DT+GNN model and record all inputs and outputs for every \texttt{UPDATE} block. Then, we train a decision tree to replace each block predicting the outputs from inputs. Thanks to the categorical states, this is a classification problem.

Unlike MLPs, decision trees cannot easily compare two features. Decision trees can compare them by building a binary tree but such trees quickly become large and hard to understand. To produce small trees, we include pairwise delta features $\Delta$, binary features for the count comparisons for every pair of states. With all MLPs replaced with decision trees, we now have a fully interpretable architecture. 

\begin{figure}[h!]
    \begin{subfigure}{0.25\textwidth}
        \centering
        \includegraphics[width=\textwidth]{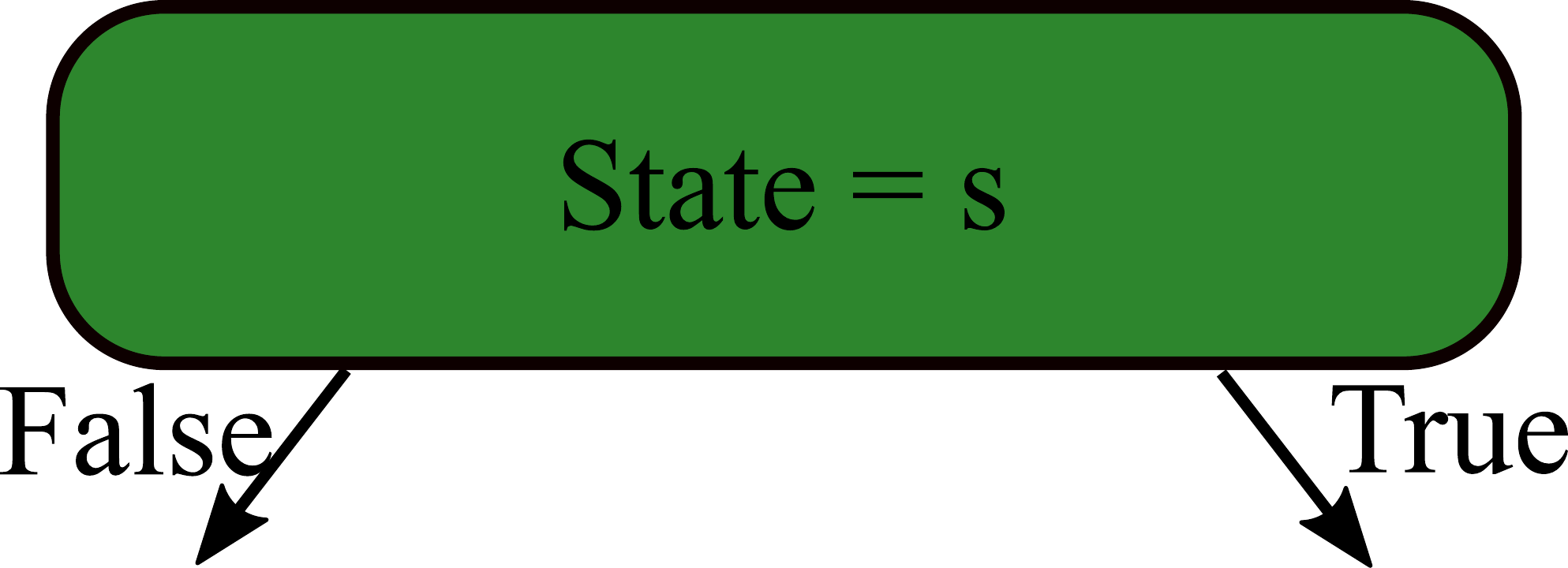}
        \caption{}
        \label{fig:state_split}
    \end{subfigure}\hfill%
    \begin{subfigure}{0.25\textwidth}
        \centering
        \includegraphics[width=\textwidth]{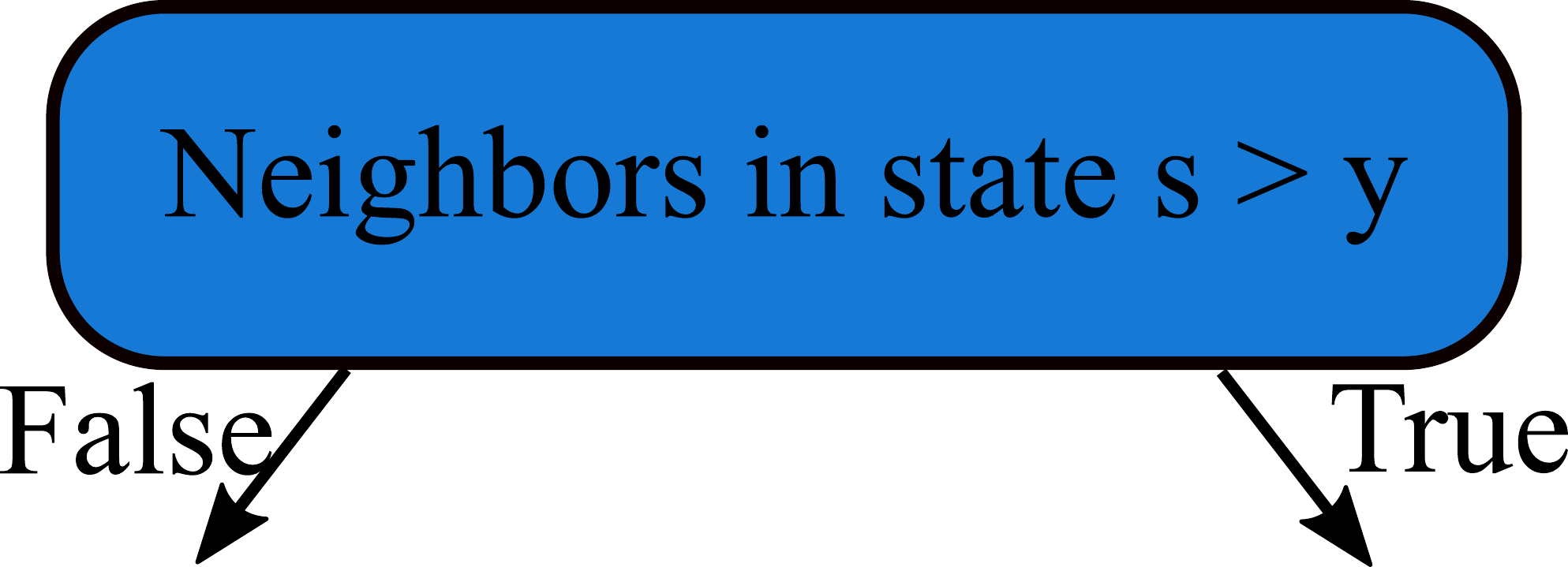}
        \caption{}
        \label{fig:neighbor_split}
    \end{subfigure}\hfill%
    \begin{subfigure}{0.25\textwidth}
        \centering
        \includegraphics[width=\textwidth]{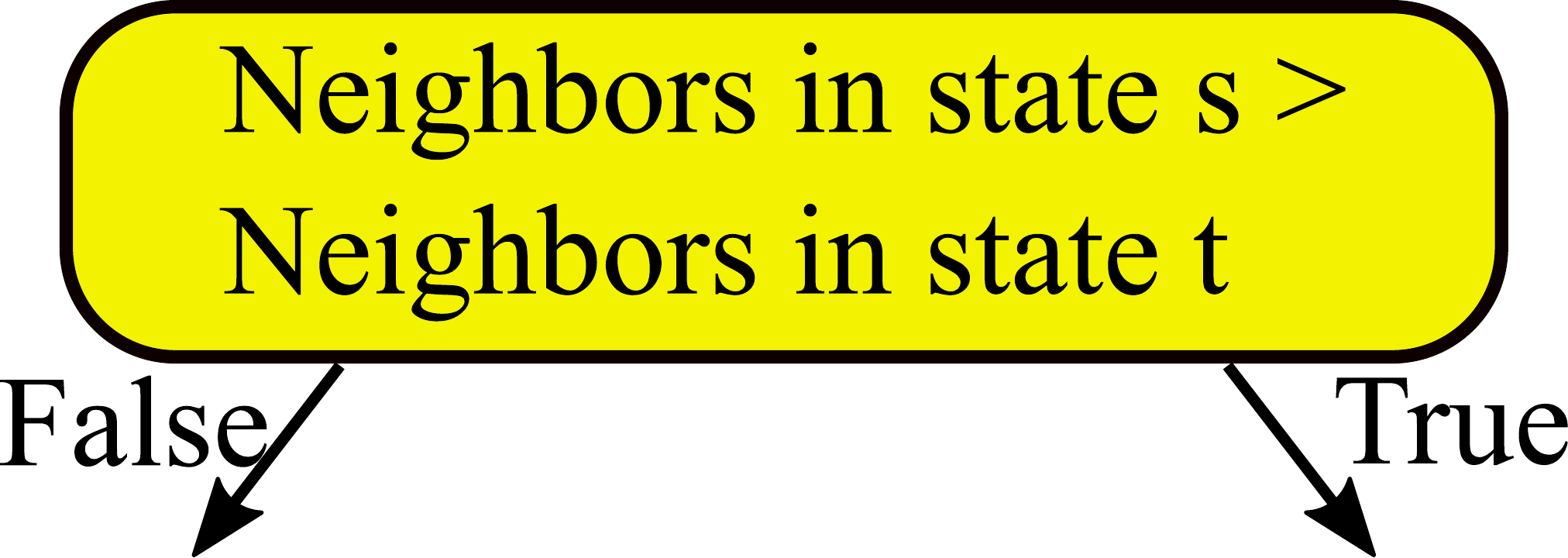}
        \caption{}
        \label{fig:delta_split}
    \end{subfigure}\hfill%
    \caption{The different branches possible in a DT+GNN layer. We can branch on (a) the state a node is in, (b) if the node has a certain number of neighbors in a certain state, or (c) if the node has more neighbors in one state than another state.}
    \label{fig:sample_decision_nodes}
\end{figure}

Figure~\ref{fig:dt_layer} shows a layer in our architecture using such decision trees. These trees define how we determine the next state for each node. The decision trees can use three different types of branching criteria for every decision node. Figure~\ref{fig:sample_decision_nodes} shows an example branch of every type.

% \colorbox{dtgreen} if we want the colors back
With the state features $\in S$, we can branch on whether or not a node is in a particular state $s$ at the beginning of the layer (Figure~\ref{fig:state_split}). Using a message feature $\in M$, a decision node branches if a node has more than $y$ neighbors in state $s$, $y$ is learnable (Figure~\ref{fig:neighbor_split}). Last, we can use branch with a delta feature $\in \Delta$ on whether a node has more neighbors in state $s$ than in state $t$ (Figure~\ref{fig:delta_split}).

\subsection{Postprocessing the DT+GNN}
\label{sec:postprocessing}

While decision trees, like MLPs, are universal function approximators if they are sufficiently deep \citep{royden1988real}, we aim to have shallow decision trees. Shallow trees are more akin to the finite state machine used in the stone age distributed computing model and are more explainable. We directly put a cap on the number of decision leaves per tree.

\paragraph{Lossless pruning.}
We prune more nodes based on the reduced error pruning algorithm~\citep{quinlan1987simplifying}. First, we define a set of data points used for pruning (pruning set). Then, we assess every decision node, starting with the one that acts upon the fewest data points. If replacing the decision node with a leaf does not lead to an accuracy drop over the entire DT+GNN on the pruning set, we replace it. We keep iterating through all decision nodes until no more changes are made.

Choosing the pruning set is not trivial. If we use the training set for pruning, all of the overfitted edge cases with few samples are used and not pruned. On the other hand, the much smaller validation set might not cover all of the decision paths and cause overpruning. Therefore, we propose to prune on both training and validation set with a different pruning criterion: A node can be pruned if replacing it with a leaf does not reduce the validation set accuracy (as in reduced error pruning) and does not reduce the training set accuracy below the validation set accuracy. Not allowing a validation accuracy drop ensures that we do not overprune. But since we allow a drop in training accuracy for the modified tree, we also remove decision nodes that result from overfitting.

\paragraph{Lossy Pruning.}
\label{sec:lossy_pruning}
Empirically, we found that we can prune substantially more decision nodes when we allow for a slight deterioration in the validation accuracy. We follow the same approach as before and always prune the decision node that leads to the smallest deterioration. We repeat this until we are satisfied with the tradeoff between the model accuracy and tree size. Defining this tradeoff is difficult as it is subjective and specific to each dataset. Therefore, we included a feature in our user interface that allows users to try different pruning steps and report the impact on accuracy. These steps are incrementally pruning $10\%$ of the nodes in the losslessly pruned decision tree.

\subsection{Generating Explanations}
\label{sec:explanation_scores}
We can use DT+GNN to create node importances as explanations similar to existing explanation methods. Formally, such an explanation $e^l_{(v,t)} \in \mathbb{R}^n$ assigns each node a real-valued importance, how much it contributed to node $v$ being in state $t$ in layer $l$. Final explanations are then taken from the final decoder layer, where states equate to classes.

We derive these explanations in a propagation procedure similar to the forward pass of a graph neural network. Initially, every node is solely responsible for its encoded state, so the initial explanation for node $v$ is $e^0_{(v,t)}$, which is a one-hot vector that is $1$ at the $v-$position, for all states $t$. But as nodes interact with their neighbors, we need to propagate explanations. Let us consider a DT+GNN layer $l$ that maps from the input state space $S$ to the target space $T$. We need to (i) understand the importance of each input feature for the decision tree, and (ii) map the importance of decision tree features back to graph nodes. For (i) we can use the Shapley values obtained by the TreeShap algorithm~\citep{lundberg2018treeshap} to assign an importance score for all features in $S, M, \Delta$ for every target state $t$. We compute these scores per node to have node-specific explanations. We can represent the TreeShap importance in three matrices $I^l_S \in \mathbb{R}^{n \times T \times S}$, $I^l_M \in \mathbb{R}^{n \times T \times S}$, $I^l_\Delta \in \mathbb{R}^{n \times T \times (S^2-S)}$. For (ii) we need to differentiate the type of feature to understand its propagation.

\paragraph{Propagating State Feature Explanations.}
Propagating explanations of one of the $S$ state features is easiest since it does not involve other nodes. The new explanation is the sum over the input state explanations, weighted by their matching feature Shapley value. There is one catch: For a state branching criterion, it can be important to \textit{not} be in a certain state, which requires inverting the explanations. We define the indicator variable $sign(s)$ which is $+1$ if node $v$ is in the input state $s$, otherwise it is $-1$. We multiply the indicator with the explanation:
\begin{equation} \label{eq:state_explanation}
    \sigma^{l+1}_{(v,t)} = \sum_{s \in S} I^l_S[v,t,s] \cdot e^l_{(v,s)} \cdot sign(s) .\nonumber
\end{equation}

\paragraph{Propagating Message Feature Explanations.}
When one of the message features is used, the node depends on the number of neighbors in the state mapping to the message feature. Let $N_s(v)$ denote the set of neighbors of node $v$ that have input state $s$. The node explanation becomes the average of the explanations of the nodes in $N_s(v)$. We can compute the next explanation by summing this expression over all input states $S$:
\begin{equation} \label{eq:message_explanation}
    \mu^{l+1}_{(v,t)} = \sum_{s \in S} I^l_M[v,t,s] \cdot \sum_{w \in N_{s}(v)} \frac{e^l_{(w,s)}}{|N_{s}(v)|} . \nonumber
\end{equation}

\paragraph{Propagating Delta Feature Explanations.}
Let us consider propagation for the feature $f(s,s')$ that is $1$ if there are more neighbors of input state $s$ than $s'$. The neighbors in $N_s(v)$ propagate explanations the same way as for a message feature. On the other hand, the nodes in $N_{s'}(v)$ work against setting $f$ to true, so we subtract their explanations. If there are more neighbors with an input state $s'$ than $s$ we have to flip the explanations of which neighbors contribute positively and negatively, for this we introduce the indicator variable $\mathbbm{1}_{>({s,s'})}$ that is either $+1$ or $-1$:
\begin{equation} \label{eq:delta_explanation}
    \delta^{l+1}_{(v,s)} = \sum_{s \in S} \sum_{s'\ne s \in S} I^l_\Delta[v, t, f(s,s')] \frac{\sum_{w \in N_{s}(v)} e^l_{(w, s)} - \sum_{w \in N_{s'}(v)} e^l_{(w, s')}}{|N_s(v)| + |N_{s'}(v)|} \cdot \mathbbm{1}_{>({s,s'})} . \nonumber
\end{equation}

The node state explanations for the next layer are the sum of these three components:
\begin{equation}
    e^{l+1}_{(v,s)} = \sigma^{l+1}_{(v,t)} + \mu^{l+1}_{(v,t)} + \delta^{l+1}_{(v,s)} . \nonumber
\end{equation}

The decoder layer is slightly special since its input features are the concatenation of node states (or node state counts for graph classification) from all other layers. For node classification, we directly compute the decoder's explanation from the respective layers. However, for graph classification the node states are pooled as follows: We supply the decoder with node state counts and count deltas as features, equivalent to the $M$ and $\Delta$ features in the DT+GNN layers. The only difference is that instead of propagating the explanation from neighbors, we now need to propagate it from all of the nodes in the graph that were in the corresponding states.

%To produce final explanations, we sum explanations for intermediate features over all layers.\\ %Depending on node or graph classification we use different intermediate features.\\
%\textbf{Node classification} uses the states of intermediate layers as intermediate features. The final explanations can be computed equivalent to Equation~\eqref{eq:state_explanation}, summed over all layers.\\
%\textbf{Graph classiciation.} The intermediate features are state counts per intermediate layer. Similar to counting neighbors, the final layers can check if counts are above a certain threshold. We also provide delta features for all pairs of counts. The explanations for the features is computed equivalent to Equation~\eqref{eq:message_explanation} and \eqref{eq:delta_explanation}. The difference that we consider \textbf{all} nodes in the state, instead of neighbors of some node.

\section{Experiments}
\subsection{Experiment setup}
\paragraph{Datasets.} We conduct experiments on two different types of datasets. First, we run DT+GNN on synthetic GNN explanation benchmarks introduced by previous literature. We use the Infection and Negative Evidence benchmarks from \citet{faber2021comparing}, The BA-Shapes, Tree-Cycle, and Tree-Grid benchmarks from \citet{ying2019gnnexplainer}, and the BA-2Motifs dataset from \citet{luo2020pgexplainer}. For all of these datasets, we know why a graph or a node should be in a particular class.  This allows us to verify that the model makes correct decisions and we compute correct explanations. Second, we experiment with the following real-world datasets: MUTAG~\citep{debnath1991structure}; BBBP~\citep{wu2017moleculenet}; Mutagenicity~\citep{kazius2005derivation}; PROTEINS, REDDIT-BINARY, IMDB-BINARY, and COLLAB~\citep{Borgwardt2005ProteinKernels}. We provide more information for all datasets, such as statistics, descriptions, and examples in Appendix~\ref{sec:datasets}. Note that all datasets are small which allows training on a few commodity CPUs. The tree construction needs little extra computation, and the cost of tree pruning is also clearly dominated by GNN training time. Precomputing all thresholds for the lossy pruning took noticeable additional time, which we sped up with a GPU for some of the larger datasets (for example, COLLAB or REDDIT-BINARY).

\paragraph{Training and Hyperparameters.}
We follow the same training scheme for all datasets following existing works~\citep{xu2018powerful}. We do a $10-$fold cross validation of the data with different splits and train both DT+GNN and a baseline GIN architecture. Both GNNs use a $2-$ layer MLP for the update function, with batch normalization~\citep{ioffe2015batch} and ReLu\citep{nair2010rectified} in between the two linear layers. GIN uses a hidden dimension of $16$, DT+GNN uses a state space of $10$. We also further divide the training set for DT+GNN to keep a holdout set for pruning decision trees. After we train Diff-DT+GNN with Gradient Descent we distill the MLPs inside into decision trees. Each tree is limited to have a maximum of $100$ nodes. The GNNs are trained on the training set for $1500$ epochs, allowing early stopping on the validation loss with a patience of $100$. Each split uses early stopping on the validation score, the results are averaged over the $10$ splits.

DT+GNN explainability allows us to see if DT+GNN uses all layers and available states. When we see unused states in the decision trees and that layers are skipped, we retrain DT+GNN with the number of layers and states that were actually used. The retrained model does not improve accuracy but it is smaller and more interpretable. We show these in Table~\ref{tab:hyperparam}. A full model is used for GIN.

\subsection{Quantitative Results}
\paragraph{DT+GNN performs comparably to GIN.}
First, we investigate the two assumptions that (i) Diff-DT+GNN matches the performance of GIN and (ii) that converting from Diff-DT+GNN to DT+GNN also comes with little loss in accuracy. We further investigate how pruning impacts DT+GNN accuracy. In Table~\ref{tab:accuracies} we report the average test set accuracy for a GIN-based GNN, Diff-DT+GNN, DT+GNN with no pruning, and the lossless version of our pruning method.

We find that DT+GNN performs almost identically to GIN. The model simplifications which increase explainability do not decrease accuracy. We observe, that tree pruning even tends to have a \textbf{positive} effect on test accuracy compared to non-pruned DT+GNN. This is likely due to the regularization induced by the pruning procedure.

\begin{table*}[ht]
\centering
\subfloat[\label{tab:accuracies}]{
\resizebox{0.673\textwidth}{!}{
\begin{tabular}{@{}l*{5}{S[table-format=-3.4]}@{}}
    \toprule
    {} & {} & \multicolumn{3}{c}{DT+GNN} \\
    \cmidrule(lr){3-5}
    {Dataset} & {GIN} & {Differentiable} & {No pruning} & {Lossless pruning}\\
    \midrule
    \makebox{Infection} & \makebox{$0.98\rpm0.04$} & \makebox{$1.00\rpm0.00$} & \makebox{$1.00\rpm0.00$} & \makebox{$1.00\rpm0.00$}\\
    \makebox{Negative} & \makebox{$1.00\rpm0.00$} & \makebox{$1.00\rpm0.00$} & \makebox{$1.00\rpm0.00$} & \makebox{$1.00\rpm0.00$}\\
    \makebox{BA-Shapes} & \makebox{$0.97\rpm0.02$} & \makebox{$1.00\rpm 0.01$} & \makebox{$0.99\rpm 0.01$} & \makebox{$0.99\rpm 0.01$}\\
    \makebox{Tree-Cycles} & \makebox{$1.00\rpm0.00$} & \makebox{$1.00\rpm0.00$} & \makebox{$1.00\rpm0.00$} & \makebox{$1.00\rpm0.00$}\\
    \makebox{Tree-Grid} & \makebox{$1.00\rpm0.01$} & \makebox{$0.99\rpm0.01$} & \makebox{$0.99\rpm0.01$} & \makebox{$0.99\rpm 0.01$}\\
    \makebox{BA-2Motifs} & \makebox{$1.00\rpm0.00$} & \makebox{$1.00\rpm0.00$} & \makebox{$1.00\rpm0.00$} & \makebox{$1.00\rpm0.00$}\\
    \midrule
    \makebox{MUTAG} & \makebox{$0.88\rpm0.05$} & \makebox{$0.88\rpm0.06$} & \makebox{$0.88\rpm 0.06$} & \makebox{$0.85\rpm 0.08$}\\
    \makebox{Mutagenicity} & \makebox{$0.81\rpm0.02$} & \makebox{$0.79\rpm0.02$} & \makebox{$0.75\rpm0.02$} & \makebox{$0.74\rpm0.02$}\\
    \makebox{BBBP} & \makebox{$0.81\rpm0.04$} & \makebox{$0.83\rpm0.03$} & \makebox{$0.82\rpm0.03$} & \makebox{$0.83\rpm0.03$}\\
    \makebox{PROTEINS} & \makebox{$0.70\rpm0.03$} & \makebox{$0.71\rpm0.02$} & \makebox{$0.71\rpm0.04$} & \makebox{$0.71\rpm0.04$}\\
    \makebox{IMDB-B} & \makebox{$0.69\rpm0.04$} & \makebox{$0.70\rpm0.05$} & \makebox{$0.69\rpm0.03$} & \makebox{$0.69\rpm0.04$}\\
    \makebox{REDDIT-B} & \makebox{$0.87\rpm0.10$} & \makebox{$0.90\rpm0.03$} & \makebox{$0.88\rpm0.03$} & \makebox{$0.87\rpm0.04$}\\
    \makebox{COLLAB} & \makebox{$0.72\rpm0.01$} & \makebox{$0.70\rpm0.02$} & \makebox{$0.69\rpm0.02$} & \makebox{$0.69\rpm0.02$}\\
\bottomrule
\end{tabular}}}
\subfloat[\label{tab:hyperparam}]{
\resizebox{0.175\textwidth}{!}{
\begin{tabular}{c*{2}{S[table-format=-3.4]}}
    \toprule
    \multicolumn{2}{c}{Hyperparameters}\\
    \cmidrule(lr){1-2}
    {Layers} & {States} \\\midrule
    \makebox{$5$} & \makebox{$6$}\\
    \makebox{$1$} & \makebox{$3$}\\
    \makebox{$5$} & \makebox{$5$}\\
    \makebox{$5$} & \makebox{$5$}\\
    \makebox{$5$} & \makebox{$5$}\\
    \makebox{$4$} & \makebox{$6$}\\
    \midrule
    \makebox{$4$} & \makebox{$6$}\\
    \makebox{$3$} & \makebox{$8$}\\
    \makebox{$3$} & \makebox{$5$}\\
    \makebox{$3$} & \makebox{$5$}\\
    \makebox{$3$} & \makebox{$5$}\\
    \makebox{$3$} & \makebox{$5$}\\
    \makebox{$3$} & \makebox{$8$}\\
\bottomrule
\end{tabular}}}
\caption{a) Test set accuracies using different GNN layers. All methods perform virtually the same for all datasets. This shows that DT+GNN layers match the expressiveness of GIN in practice. (b) DT+GNN hyperparameters found through tree inspection.} 
\end{table*}

\paragraph{Pruning significantly reduces the decision tree sizes.}
Second, we examine the effectiveness of our pruning method. We compare the tree sizes before pruning, after lossless pruning, and after lossy pruning. We measure tree size as the sum of decision nodes over all trees. Additionally, we verify the effectiveness of using our pruning criterion for reduced error pruning and compare it against simpler setups of using only the training or validation set for pruning. We report the tree sizes and test set accuracies for all configurations in Table~\ref{tab:pruning}.

We can see that the reduced error pruning leads to an impressive drop in the number of nodes required in the decision trees. On average, we can prune about $62\%$ of nodes in synthetic datasets and even around $84\%$ of nodes in real-world datasets without a loss in accuracy. If we accept small drops, in accuracy we can even a total of $68\%$ and $87\%$ of nodes in synthetic and real-world datasets, respectively. If we compare the different approaches for reduced error pruning, we can see that our proposed approach of using both training and validation accuracy performs the best. As expected, pruning only on the validation set tends to overprune the trees: Trees become even smaller but there is also a larger drop in accuracy, especially in the real-world datasets. Using the training set tends to underprune, there is no drop in accuracy but decision trees for real-world graphs tend to stay large.

\begin{table*}[ht]
\centering
\resizebox{\textwidth}{!}{
\begin{tabular}{@{}l*{11}{S[table-format=-3.4]}@{}}
    \toprule
    %& & & %\multicolumn{6}{c}{Reduced Error Pruning} & &\\
    %\cmidrule(lr){4-9} 
    & \multicolumn{2}{c}{No pruning} & \multicolumn{2}{c}{REP Training} & \multicolumn{2}{c}{REP Validation} & \multicolumn{2}{c}{REP Ours} & \multicolumn{2}{c}{REP Lossy}\\
    \cmidrule(lr){2-3} \cmidrule(lr){4-5} \cmidrule(lr){6-7} \cmidrule(lr){8-9} \cmidrule(lr){10-11}
    {Dataset} & {Accuracy} & {Size} & {Accuracy} & {Size} & {Accuracy} & {Size} & {Accuracy} & {Size} & {Accuracy} & {Size} \\
    \midrule
    \makebox{Infection} & \makebox{$1.00\rpm0.00$} & \makebox{$205\rpm56$} & \makebox{$1.00\rpm0.00$} & \makebox{$26\rpm2$} & \makebox{$1.00\rpm0.00$} & \makebox{$25\rpm2$} & \makebox{$1.00\rpm0.00$} & \makebox{$26\rpm2$} & \makebox{$0.98\rpm0.01$} & \makebox{$17\rpm2$}\\
    \makebox{Negative} & \makebox{$1.00\rpm0.00$} & \makebox{$18\rpm14$} & \makebox{$1.00\rpm0.00$} & \makebox{$5\rpm0$} & \makebox{$1.00\rpm0.00$} & \makebox{$5\rpm0$} & \makebox{$1.00\rpm0.00$} & \makebox{$5\rpm0$} & \makebox{$1.00\rpm0.00$} & \makebox{$4\rpm0$}\\
    \makebox{BA-Shapes} & \makebox{$0.99\rpm0.01$} & \makebox{$30\rpm10$} & \makebox{$0.99\rpm0.01$} & \makebox{$21\rpm5$} & \makebox{$0.97\rpm0.03$} & \makebox{$15\rpm4$} & \makebox{$0.99\rpm0.01$} & \makebox{$21\rpm5$} & \makebox{$0.98\rpm0.04$} & \makebox{$17\rpm4$}\\
    \makebox{Tree-Cycles} & \makebox{$1.00\rpm0.00$} & \makebox{$19\rpm5$} & \makebox{$1.00\rpm0.00$} & \makebox{$11\rpm3$} & \makebox{$0.99\rpm0.02$} & \makebox{$9\rpm2$} & \makebox{$1.00\rpm0.00$} & \makebox{$11\rpm3$} & \makebox{$0.99\rpm0.01$} & \makebox{$9\rpm3$}\\
    \makebox{Tree-Grid} & \makebox{$0.99\rpm0.01$} & \makebox{$30\rpm13$} & \makebox{$0.99\rpm0.01$} & \makebox{$17\rpm8$} & \makebox{$0.99\rpm0.01$} & \makebox{$13\rpm4$} & \makebox{$0.99\rpm0.01$} & \makebox{$15\rpm8$} & \makebox{$0.99\rpm0.01$} & \makebox{$15\rpm8$}\\
    \makebox{BA-2Motifs} & \makebox{$1.00\rpm0.00$} & \makebox{$141\rpm43$} & \makebox{$1.00\rpm0.00$} & \makebox{$12\rpm3$} & \makebox{$1.00\rpm0.01$} & \makebox{$11\rpm3$} & \makebox{$1.00\rpm0.00$} & \makebox{$13\rpm4$} & \makebox{$1.00\rpm0.00$} & \makebox{$13\rpm4$}\\
    \midrule
    \makebox{MUTAG} & \makebox{$0.88\rpm0.06$} & \makebox{$59\rpm27$} & \makebox{$0.86\rpm0.08$} & \makebox{$19\rpm17$} & \makebox{$0.83\rpm0.07$} & \makebox{$7\rpm6$} & \makebox{$0.85\rpm0.08$} & \makebox{$18\rpm16$} & \makebox{$0.85\rpm0.08$} & \makebox{$18\rpm16$}\\
    \makebox{Mutagenicity} & \makebox{$0.75\rpm0.02$} & \makebox{$375\rpm13$} & \makebox{$0.76\rpm0.02$} & \makebox{$154\rpm19$} & \makebox{$0.73\rpm0.01$} & \makebox{$56\rpm16$} & \makebox{$0.74\rpm0.02$} & \makebox{$91\rpm36$} & \makebox{$0.73\rpm0.02$} & \makebox{$50\rpm19$}\\
    \makebox{BBBP} & \makebox{$0.82\rpm0.03$} & \makebox{$366\rpm53$} & \makebox{$0.84\rpm0.02$} & \makebox{$88\rpm52$} & \makebox{$0.79\rpm0.04$} & \makebox{$8\rpm10$} & \makebox{$0.83\rpm0.03$} & \makebox{$46\rpm27$} & \makebox{$0.82\rpm0.03$} & \makebox{$31\rpm18$}\\
    \makebox{PROTEINS} & \makebox{$0.71\rpm0.04$} & \makebox{$206\rpm90$} & \makebox{$0.72\rpm0.03$} & \makebox{$12\rpm13$} & \makebox{$0.70\rpm0.04$} & \makebox{$8\rpm6$} & \makebox{$0.71\rpm0.04$} & \makebox{$9\rpm6$} & \makebox{$0.71\rpm0.04$} & \makebox{$9\rpm6$}\\
    \makebox{IMDB-B} & \makebox{$0.69\rpm0.03$} & \makebox{$218\rpm32$} & \makebox{$0.69\rpm0.04$} & \makebox{$20\rpm9$} & \makebox{$0.66\rpm0.06$} & \makebox{$16\rpm6$} & \makebox{$0.69\rpm0.04$} & \makebox{$29\rpm9$} & \makebox{$0.69\rpm0.04$} & \makebox{$29\rpm9$}\\
    \makebox{REDDIT-B} & \makebox{$0.88\rpm0.03$} & \makebox{$248\rpm28$} & \makebox{$0.88\rpm0.02$} & \makebox{$53\rpm14$} & \makebox{$0.85\rpm0.04$} & \makebox{$28\rpm8$} & \makebox{$0.87\rpm0.04$} & \makebox{$49\rpm21$} & \makebox{$0.87\rpm0.04$} & \makebox{$38\rpm15$}\\
    \makebox{COLLAB} & \makebox{$0.69\rpm0.02$} & \makebox{$301\rpm1$} & \makebox{$0.70\rpm0.02$} & \makebox{$36\rpm15$} & \makebox{$0.67\rpm0.03$} & \makebox{$22\rpm12$} & \makebox{$0.69\rpm0.02$} & \makebox{$30\rpm18$} & \makebox{$0.68\rpm0.02$} & \makebox{$21\rpm12$}\\
\bottomrule
\end{tabular}}
\caption{Running reduced error pruning (REP) on different pruning sets. Lossy prunes the nodes with least loss in accuracy up to a manually chosen threshold.}
\label{tab:pruning}
\end{table*}  

\subsection{Qualitative Results}
\label{sec:example_interpretations}
\begin{figure}
    \begin{subfigure}[t]{0.25\textwidth}
        \centering
        \includegraphics[width=\textwidth]{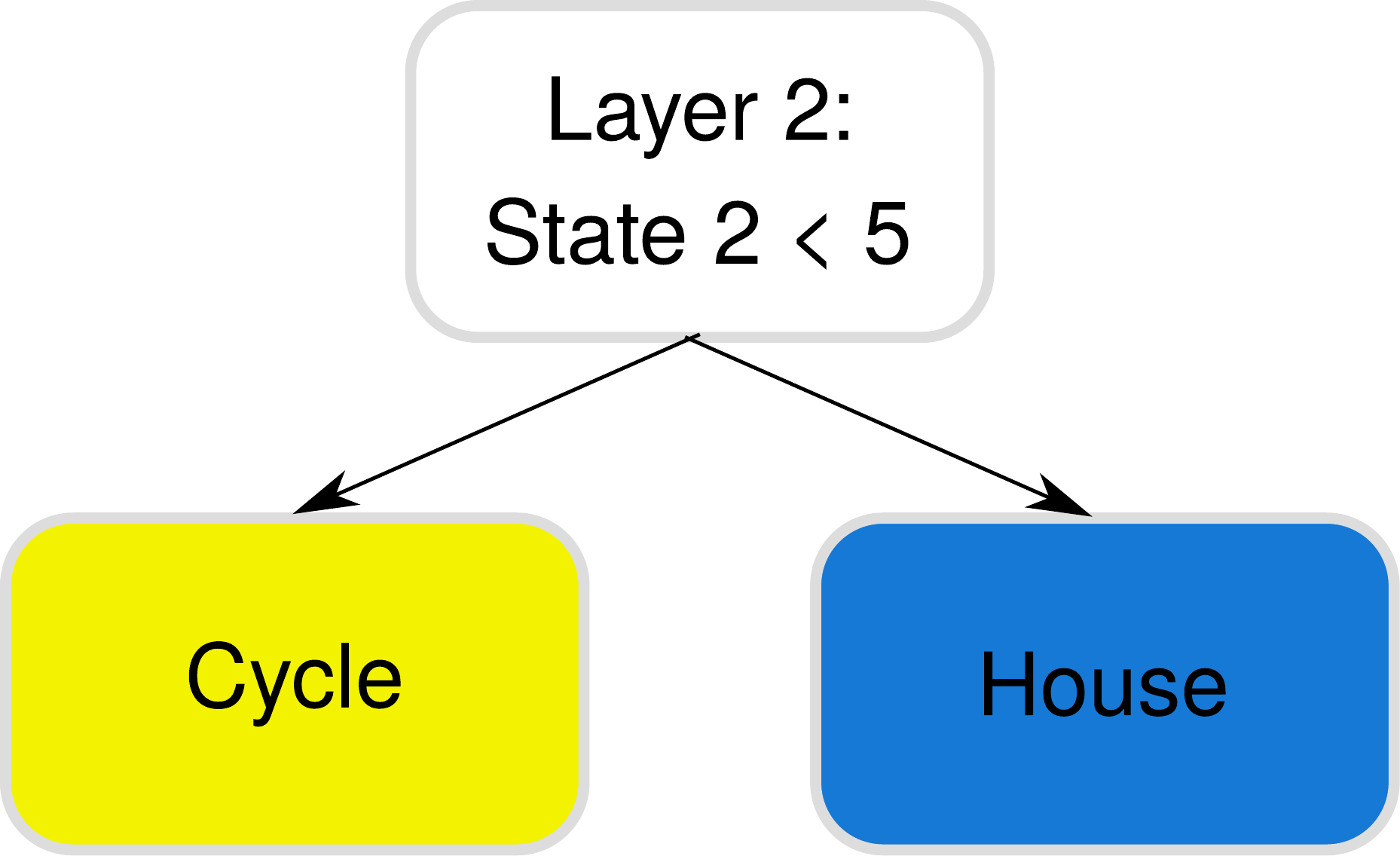}
        \caption{}
        \label{fig:bias_example_output}
    \end{subfigure}\hfill%
    \begin{subfigure}[t]{0.28\textwidth}
        \centering
        \includegraphics[width=\textwidth]{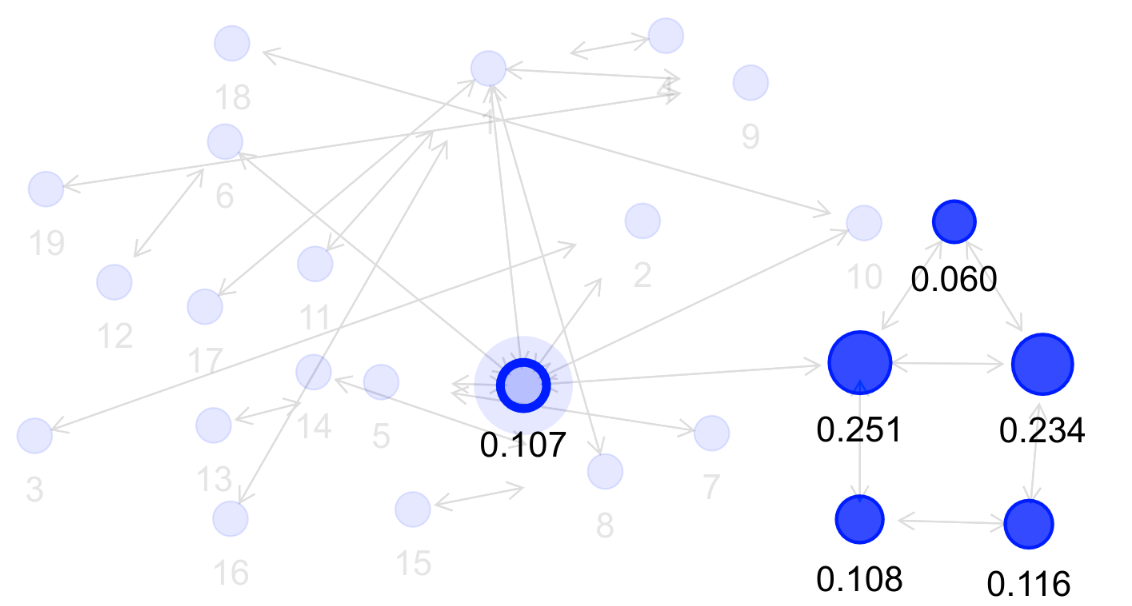}
        \caption{}
        \label{fig:bias_example_house}
    \end{subfigure}\hfill%
    \begin{subfigure}[t]{0.28\textwidth}
        \centering
        \includegraphics[width=\textwidth]{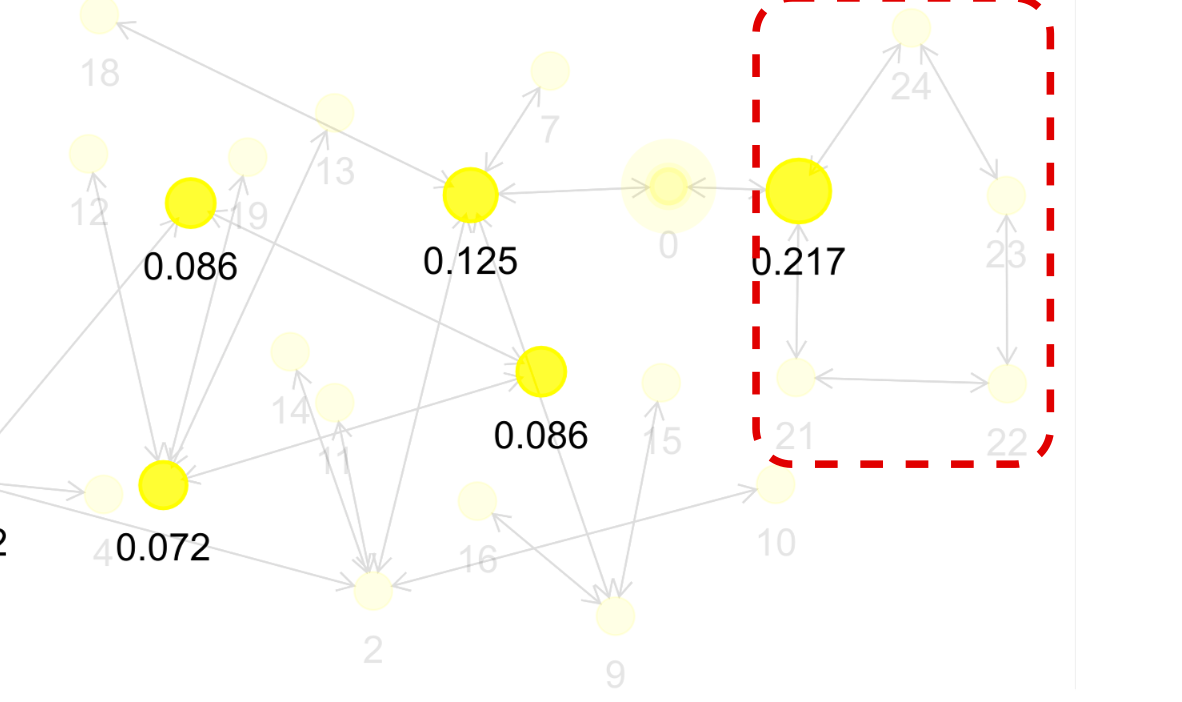}
        \caption{}
        \label{fig:bias_example_cycle}
    \end{subfigure}\hfill%
    \caption{DT+GNN uses bias terms to solve the BA-2Motifs dataset. (a) The decoder layer only learns what a house is, the cycle is ``not'' a house (b) and (c) Explanation graphs: numbers denote the importance of each node, non-important nodes are transparent. (b) Correct explanation for the house on the right since all nodes in the house receive a high importance. (c) Incorrect explanation for the cycle in the the dashed red box since only one node in the cycle receives a high importance.}
\end{figure}

\textbf{Bias Terms.}
Bias terms are often overlooked but can be problematic for importance explanations~\citep{wang2019bias}. When a GNN uses bias terms to predict a class, nothing of the input was used so nothing should be important~\citep{faber2021comparing}. We observe this in the BA-2Motifs dataset. In this dataset, a GNN needs to predict if a given graph contains a house structure or a cycle. Figure~\ref{fig:bias_example_output} shows the output layer of DT+GNN for this dataset. If at least $5$ nodes are in the house (in state $2$ in layer $2$), the graph is classified as house, otherwise, it is a cycle. DT+GNN only learns what a house is, cycles are then ``not houses''. Consequently, the explanations for which nodes contribute to the classification as a house are correct (Figure~\ref{fig:bias_example_house}) but the explanations for cycles are not (Figure~\ref{fig:bias_example_cycle}). 

\label{fig:bias_example}

\begin{figure}[h!]
    \begin{subfigure}{0.25\textwidth}
        \centering
        \includegraphics[width=\textwidth]{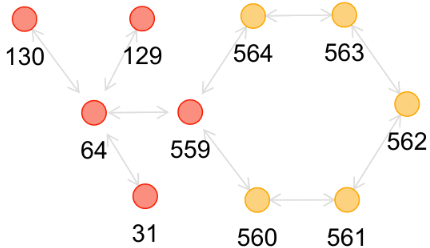}
        \caption{}
        \label{fig:cycle_layer1}
    \end{subfigure}\hfill%
    \begin{subfigure}{0.25\textwidth}
        \centering
        \includegraphics[width=\textwidth]{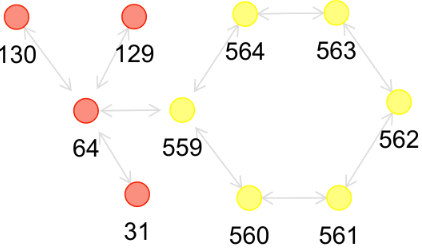}
        \caption{}
        \label{fig:cycle_layer2}
    \end{subfigure}\hfill%
    \begin{subfigure}{0.25\textwidth}
        \centering
        \includegraphics[width=\textwidth]{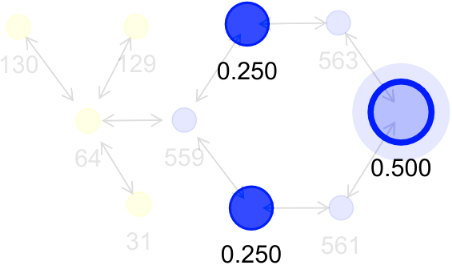}
        \caption{}
        \label{fig:cycle_explanations}
    \end{subfigure}\hfill%
    \caption{DT+GNN solves an instance of the Tree-Cycle dataset. (a) and (b) DT+GNN learns states for all nodes. Numbers are node IDs, different colors map to states. (a) DT+GNN finds nodes with a degree of $2$. (b) DT+GNN finds nodes connected to degree $2$ nodes, this already distinguishes the cycle from the remaining graph. (c) Explanations for the rightmost node, high numbers show nodes which important for that node's final state. The correct explanation are the neighbors of the neighbors.}
\end{figure}

\textbf{Surplus ground truth.}
\citet{faber2021comparing} discuss that having more ground truth evidence than necessary can also cause problems with explanations. For example, we observe this problem on the Tree-Cycle dataset. DT+GNN uses the second layer for making the final prediction. At this point, no node could even see the whole cycle. This is because the base graph is a balanced binary tree. Apart from the root node, no nodes other than cycle nodes have degree $2$. In principle, DT+GNN learns to predict cycle nodes as those nodes having degree $2$ neighbors (Figure~\ref{fig:cycle_layer1} and \ref{fig:cycle_layer2}). For a node $n$, DT+GNN assigns explanations to the nodes causing $n$ to have degree $2$ neighbors. Figure~\ref{fig:cycle_explanations} shows an example explanation. The explanation for the highlighted node is the neighbors of its neighbors, the other cycle nodes are unnecessary. Therefore, they are not and should not be part of the explanation.

\label{fig:treecycle_example}

\textbf{Simple solution for MUTAG} We find some inconsistencies regarding the MUTAG and Mutagenicity datasets. There are several works~\citep{duval2021graphsvx, luo2020pgexplainer, ying2019gnnexplainer, yuan2020explainabilitySurvey, yuan2021subgraphx} that use the Mutagenicity dataset but call it MUTAG. This can lead to errors when finding explanations. Previous explanation methods\citep{duval2021graphsvx, faber2020contrastive, ying2019gnnexplainer, yuan2021subgraphx} use the existence of $NO_2$ subgraphs as correct explanations~\citep{debnath1991structure}. However, this explanation is not correct in MUTAG since all graphs have this group. For MUTAG, we show a simple solution based on degree counting with $89\%$ accuracy in Appendix~\ref{sec:more_example_interpretations}.

%\textbf{Simple solution for PROTEINS} We observe that our decoder layer for DT+GNN only uses features from the input layer. This raises the question if we even need any message passing at all on this dataset? Indeed, we find that simply feeding every node through a $2-$layer MLP and applying sum-pooling and readout MLP achieves $76\rpm3\%$ accuracy, which is competitive with GNNs~\citep{papp2021dropgnn, errica2020fair}.

\section{Conclusion}
In this paper, we presented DT+GNN which is a fully explainable graph learning method. Full explainability means that we can follow the decision process of DT+GNN and observe how information is used in every layer, yielding an inherently explainable and understandable model. Under the hood, DT+GNN employs decision trees to achieve its explainability. We experimentally verify that the slightly weaker GNN layers used do not have a large negative impact on the accuracy and that the employed tree pruning methods are very effective. We also provide some examples of how DT+GNN can help to gain insights into different GNN problems. Moreover, we also provide a user interface that allows easy and interactive exploration with DT+GNN. As a limitation, we have observed, that in datasets that have many node input features such as Cora~\citep{sen2008collective} or OGB-ArXiv~\citep{hu2020ogb}, it is hard to successfully fit a decision tree to the MLP that embeds those input features. Future work could tackle this, for example, by using PCA, clustering, or special MLP construction techniques~\citep{wu2017beyond,schaaf2019enhancing}.

\paragraph{Impact statement.} We believe that DT+GNN will help improve our understanding of GNNs and graph learning tasks they are used for. We hope that this leads to increased transparency of predictions made by GNNs. This will be crucial in the adoption of GNNs in more critical domains such as medicine and should help avoid models that make biased or discriminatory decisions. Similar to the MUTAG or PROTEINS examples, this transparency can also help experts in various domains to better understand their datasets and improve their approaches.

% We believe that DT+GNN makes a big step towards better understanding of graph machine learning and GNNs. The impact we hope to achieve is to increase the transparency of predictions done with this kind of GNNs. Transparency might be crucial to adoption of GNNs in more critical domains. Additionally, transparency can help to understand if a model does biased or discriminating decisions. Another domain where we hope DT+GNN can make a positive impact is prodiving insights into datasets for humans. Similar to the insights we got for MUTAG or REDDIT-BINARY, we hope that experts in more difficult and important domains can understand their dataset better and hopefully inspire them for improving their approaches.

\iffalse
\begin{ack}
Use unnumbered first level headings for the acknowledgments. All acknowledgments
go at the end of the paper before the list of references. Moreover, you are required to declare
funding (financial activities supporting the submitted work) and competing interests (related financial activities outside the submitted work).
More information about this disclosure can be found at: \url{https://neurips.cc/Conferences/2022/PaperInformation/FundingDisclosure}.

Do {\bf not} include this section in the anonymized submission, only in the final paper. You can use the \texttt{ack} environment provided in the style file to autmoatically hide this section in the anonymized submission.
\end{ack}
\fi

\bibliography{paper}
\bibliographystyle{abbrvnat}

\appendix

\clearpage
\section{Using the tool}
\label{sec:manual}
\begin{figure}[bh]
    \centering
    \includegraphics[width=\textwidth]{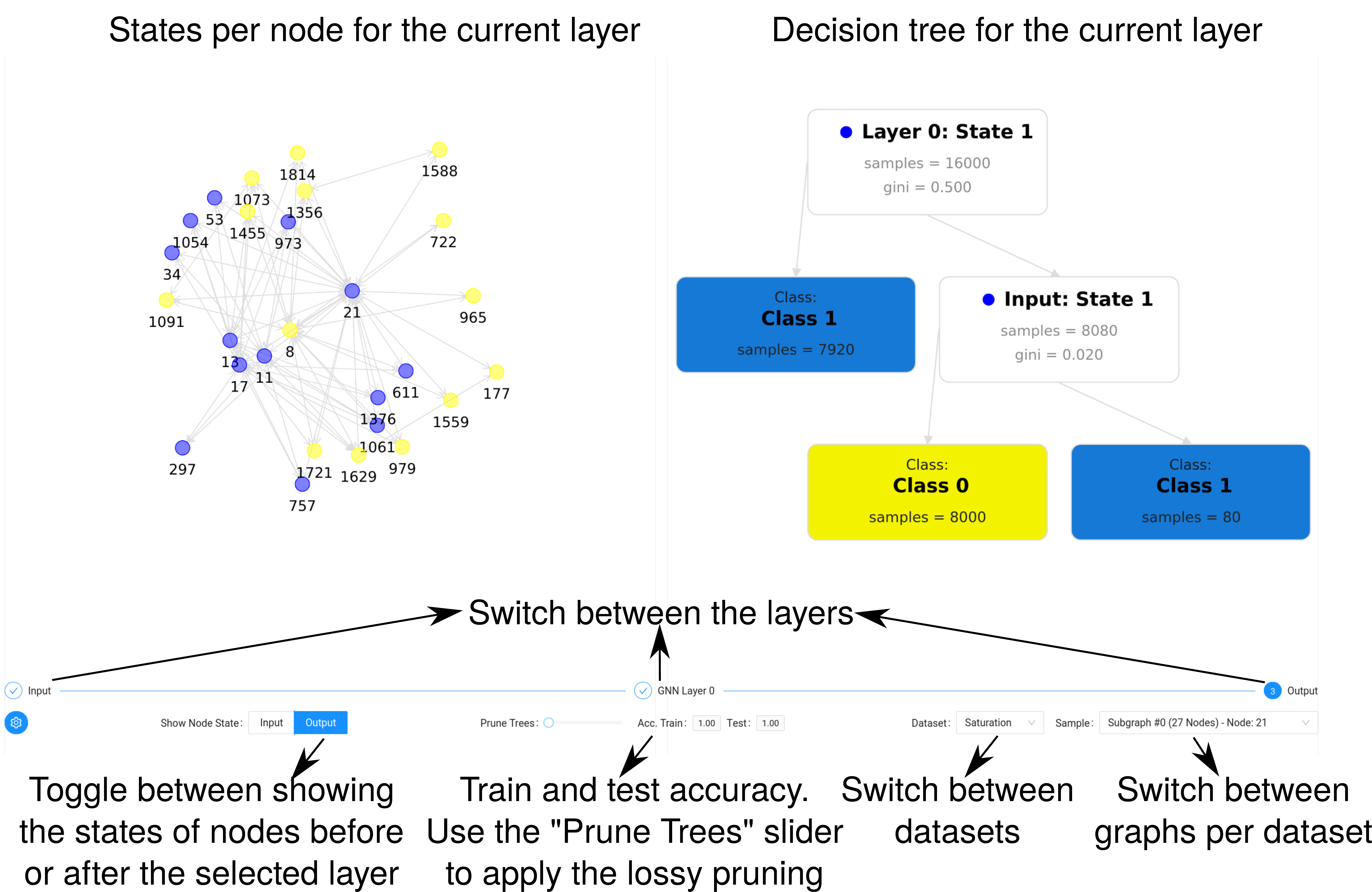}
    \caption{Initial page for the web tool. We can see the decision trees for DT+GNN per dataset and which node for a graph is in what state. We can switch layers, graphs and datasets. We can also see the test accuracy for the current setting and choose an amount of lossy pruning. with the slider.}
    \label{fig:tool_startpage}
\end{figure}

A example instance of the tool is deployed and available via Netlify\footnote{\url{https://netlify.com/}} and can be accessed under the link \url{https://interpretable-gnn.netlify.app/}. The supplementary material also contains code to host the interface yourself, in case you want to try variations of DT+GNN. In the backend, we use PyTorch~\citep{paszke2019pytorch}\footnote{\url{https://github.com/pytorch/pytorch}} and PyTorch Geometric~\citep{fey2019fast}\footnote{\url{https://github.com/pyg-team/pytorch_geometric}} to train DT+GNN and SKLearn\citep{pedregosa2011scikit}\footnote{\url{https://github.com/scikit-learn/scikit-learn}} to train decision trees. 

The tool is built with React, in particular the Ant Design library.\footnote{\url{https://github.com/ant-design/ant-design/}} We visualize graphs with the Graphin library.\footnote{\url{https://github.com/antvis/Graphin}} The interface is a single page that will look similar to Figure~\ref{fig:tool_startpage}.

The largest part of the interface is taken by two different panels at the top. In the right panel, you can see the decision tree for the currently selected layer. The trees use the three branching options from Figure~\ref{fig:sample_decision_nodes}. In the interface, evaluating the branching to true means taking the left path (this is opposite to Figure~\ref{fig:sample_decision_nodes}, which we will flip). In the left panel, you can see an example graph and which nodes end up in which state after this layer (in the bottom left you can toggle to see the input states instead). This panel does not show the full graph (most graphs in the datasets are prohibitively large) but an excerpt around an interesting region. Directly below these two graphics, you have the option to switch between layers by clicking on the respective bubble.

In the bottom right, you can switch to a different graph in the same dataset or to a different dataset. In the centre, you can see the accuracy of DT+GNN with the displayed layers. The slider allows to apply the lossy pruning from Section~\ref{sec:lossy_pruning} and the accuracy values update to the selected pruning level.

\begin{figure}[bh]
    \centering
    \includegraphics[width=\textwidth]{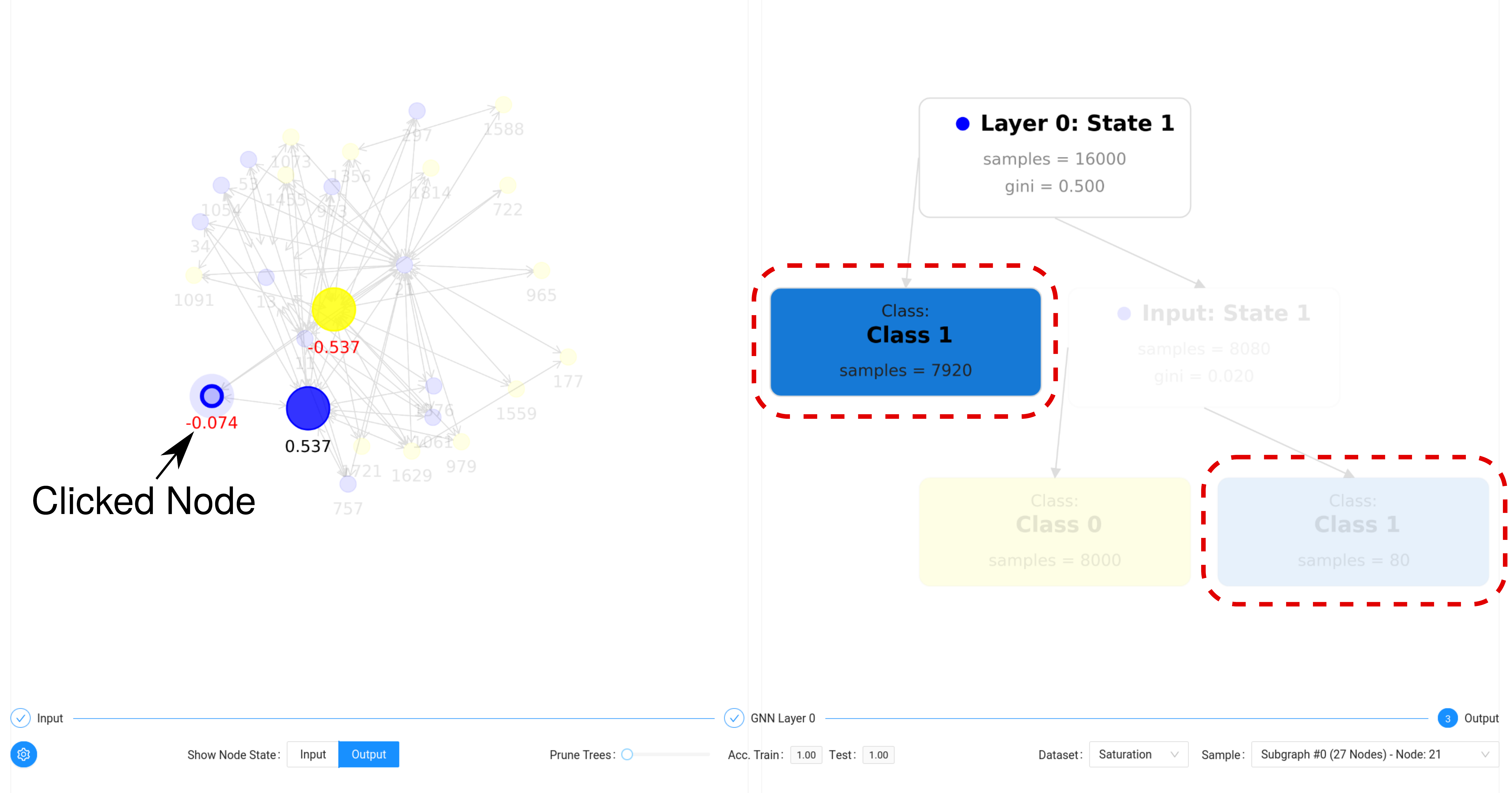}
    \caption{Interface when clicking on a node for closer examination. We can see node-level importance scores for this node on the left and the taken decision path on the right. Two paths end in the blue state, shown by the red boxes. The path the node takes is highlighted, the other path is blurred out.}
    \label{fig:tool_select}
\end{figure}

The interface also allows us to examine a single node more closely by clicking on it (see Figure~\ref{fig:tool_select}; here we clicked the blue node on the very right). Selecting reveals two things: In the graph panel, you can see the explanation scores from Section~\ref{sec:explanation_scores} for this node in this layer. In the tree panel, you can see the decision path in the tree for this node. This is particularly helpful if multiple leaves in the tree would lead to the same output state as in this example.

\clearpage
\section{Datasets}
\label{sec:datasets}
\subsection{Synthetic Datasets}
\begin{itemize}
    \item \textbf{Infection} \cite{faber2021comparing} is a synthetic node classification dataset. This dataset consists of randomly generated directed graphs, where each node can be healthy or infected. The classification task predicts the length of the shortest directed path from an infected node.
    
    \item \textbf{Negative Evidence} \cite{faber2021comparing} is a synthetic node classification dataset. A random graph with ten red nodes, ten blue nodes, and 1980 white nodes is created. The task is to determine whether the white nodes have more red or blue neighbours.
    
    \item \textbf{BA Shapes} \cite{ying2019gnnexplainer} is a synthetic node classification dataset. Each graph contains a Barabasi-Albert (BA) base graph and several house-like motifs attached to random nodes of the base graph. The node labels are determined by the node's position in the house motif or base graph.
    
    \item \textbf{Tree Cycle} \cite{ying2019gnnexplainer} is a synthetic node classification dataset. Each graph contains an 8-level balanced binary tree and a six-node cycle motif attached to random nodes of the tree. The classification task predicts whether the nodes are part of the motif or tree.
    
    \item \textbf{Tree Grid} \cite{ying2019gnnexplainer} is a synthetic node classification dataset. Each graph contains an 8-level balanced binary tree and a 3-by-3 grid motif attached to random nodes of the tree. The classification task predicts whether the nodes are part of the motif or the tree.
        
    \item \textbf{BA 2Motifs} \cite{luo2020pgexplainer} is a synthetic graph classification dataset. Barabasi-Albert graphs are used as the base graph. Half of the graphs have a house-like motif attached to a random node, and the other half have a five-node cycle. The prediction task is to classify each graph, whether it contains a house or a cycle.
    
\end{itemize}

\newcommand\sampleWidth{0.22}
\newcommand\sampleSpacing{\hspace{0.5em}} 

\begin{figure}[htb]
\begin{center}
\centering
    \subfloat[\centering Infection Input]{{\includegraphics[width=\sampleWidth\textwidth]{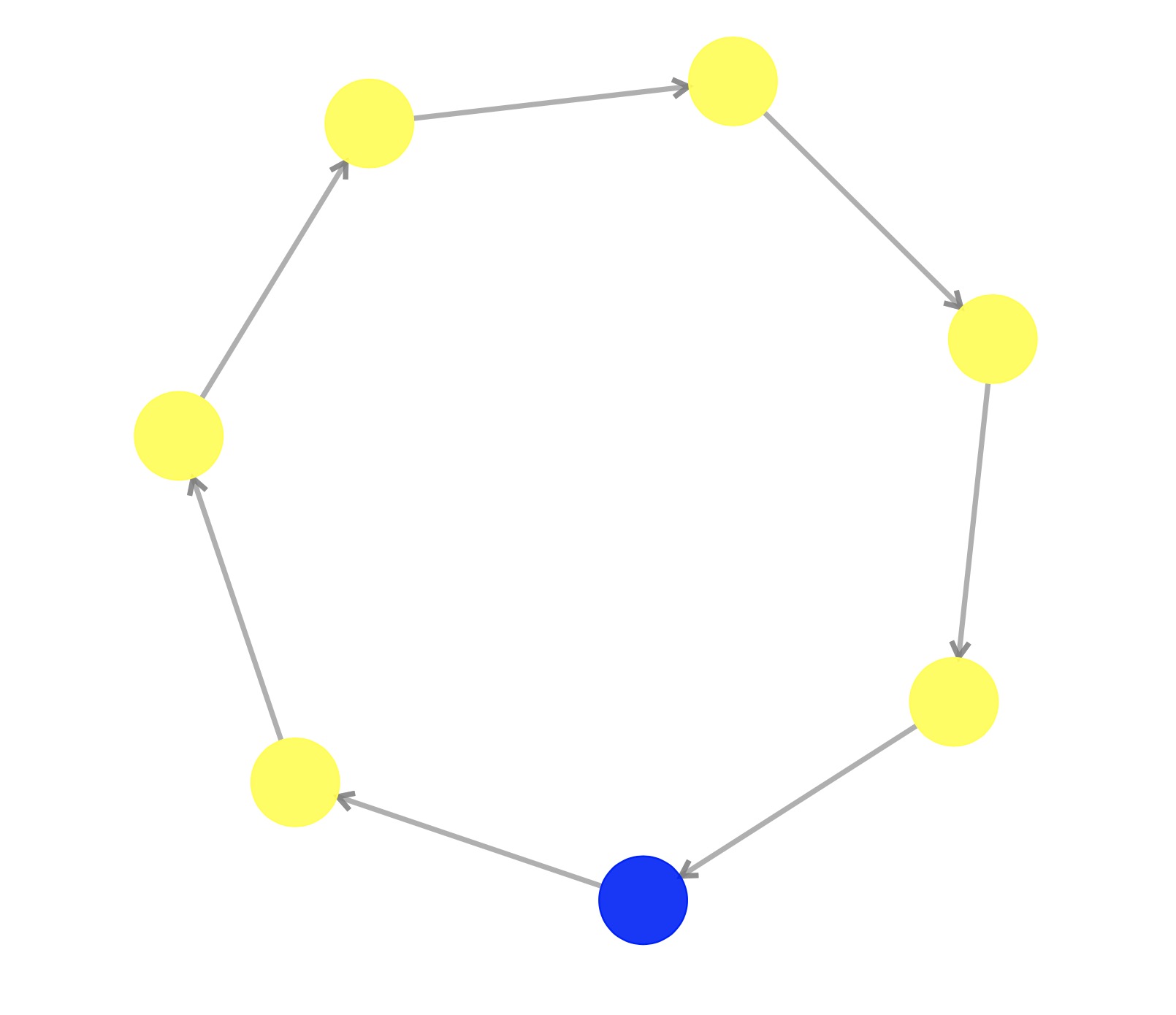} }}
    \sampleSpacing
    \subfloat[\centering Infection Prediciton]{{\includegraphics[width=\sampleWidth\textwidth]{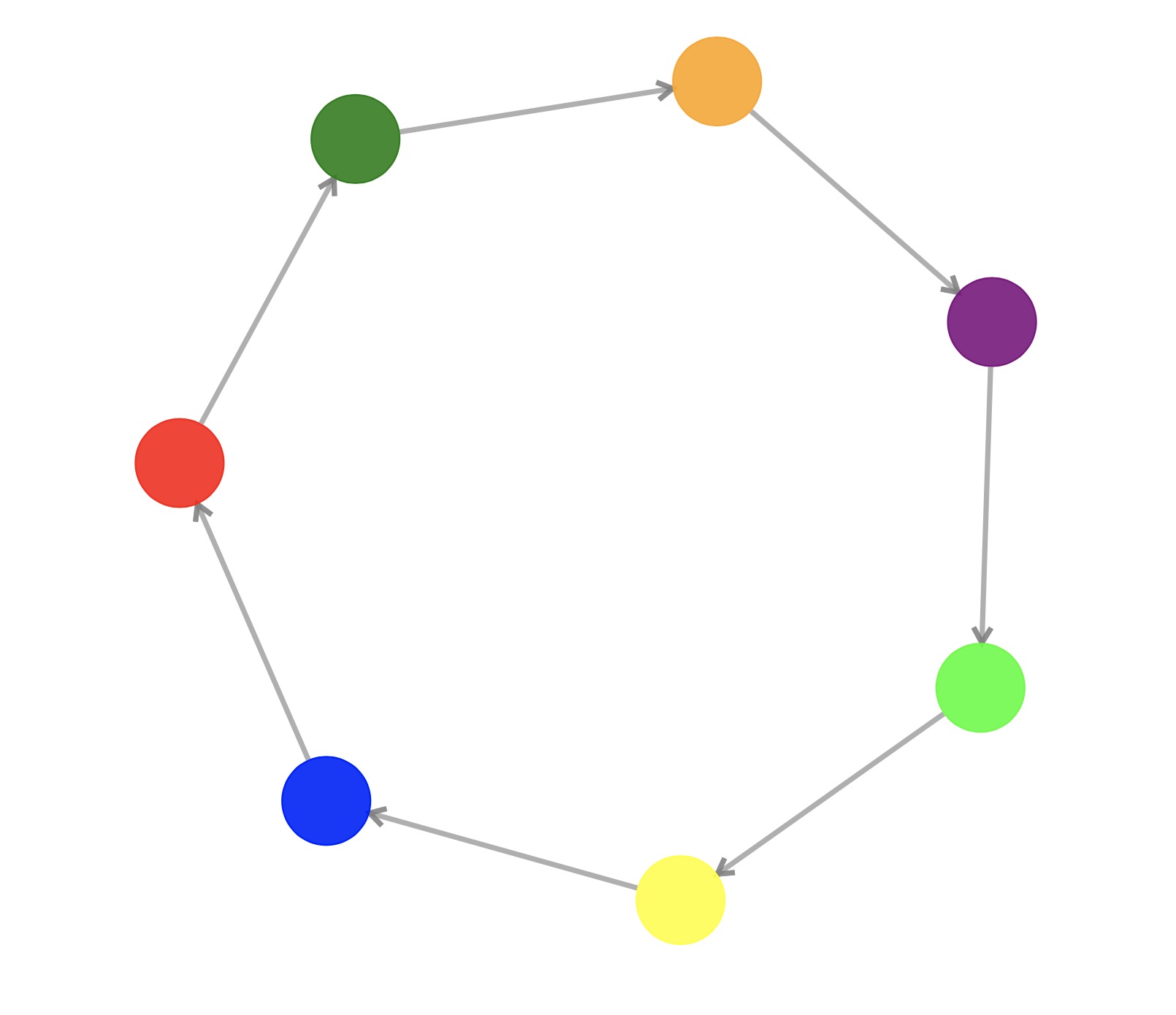} }}
    \sampleSpacing
    \subfloat[\centering Saturation Input]{{\includegraphics[width=\sampleWidth\textwidth]{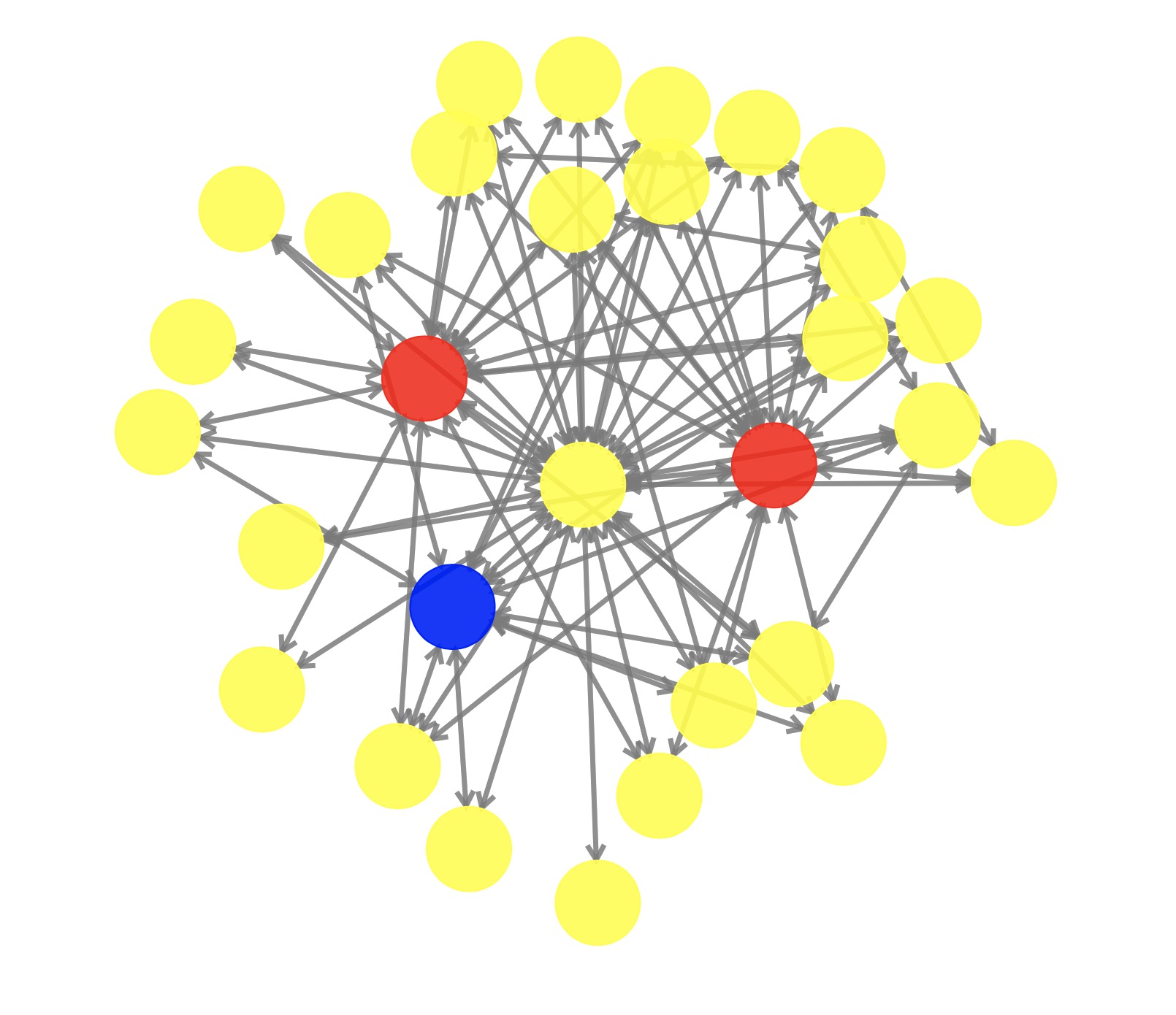} }}
    \sampleSpacing
    \subfloat[\centering Saturation Prediciton]{{\includegraphics[width=\sampleWidth\textwidth]{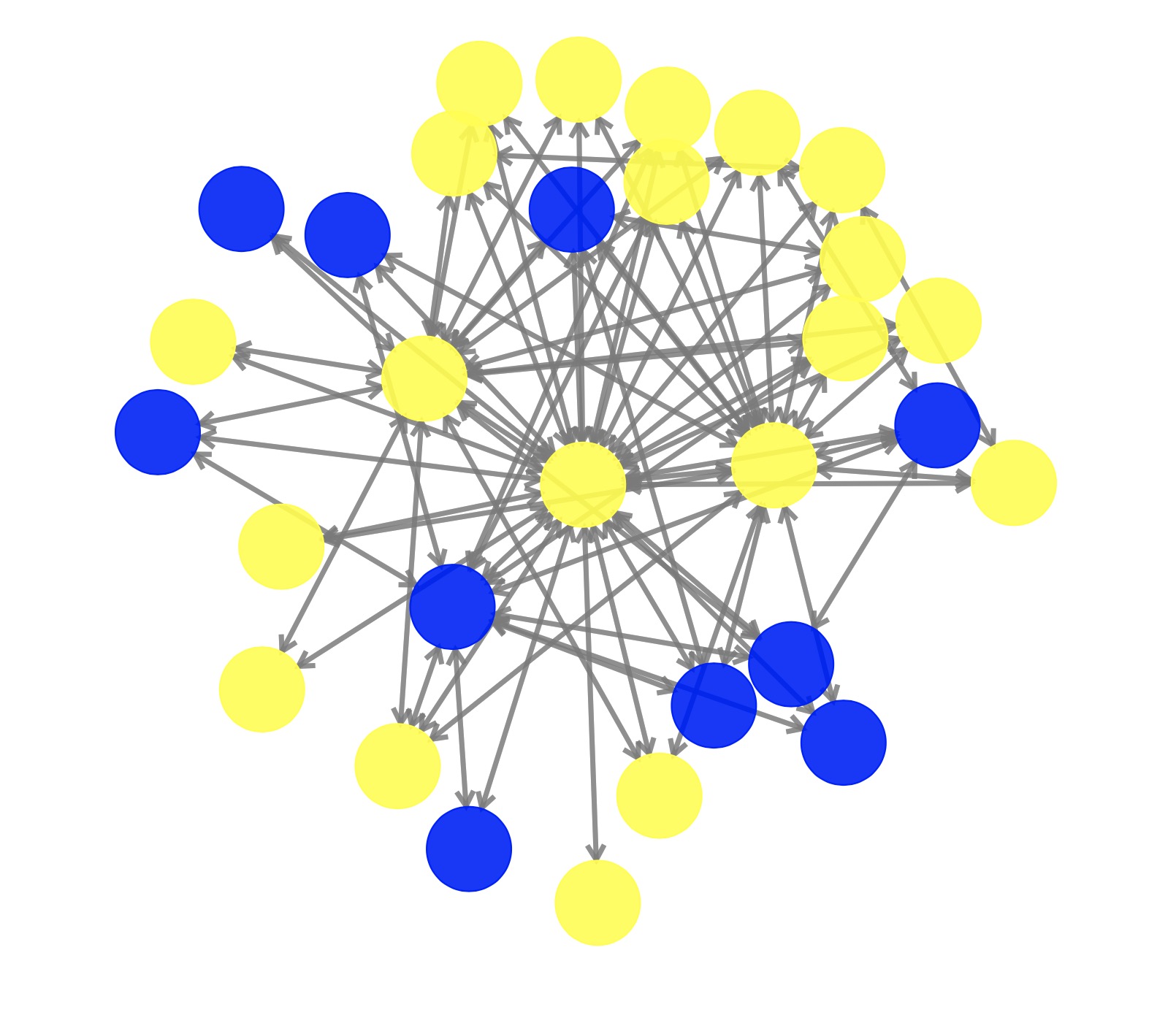} }}
    \sampleSpacing
    \subfloat[\centering BA Shapes Input]{{\includegraphics[width=\sampleWidth\textwidth]{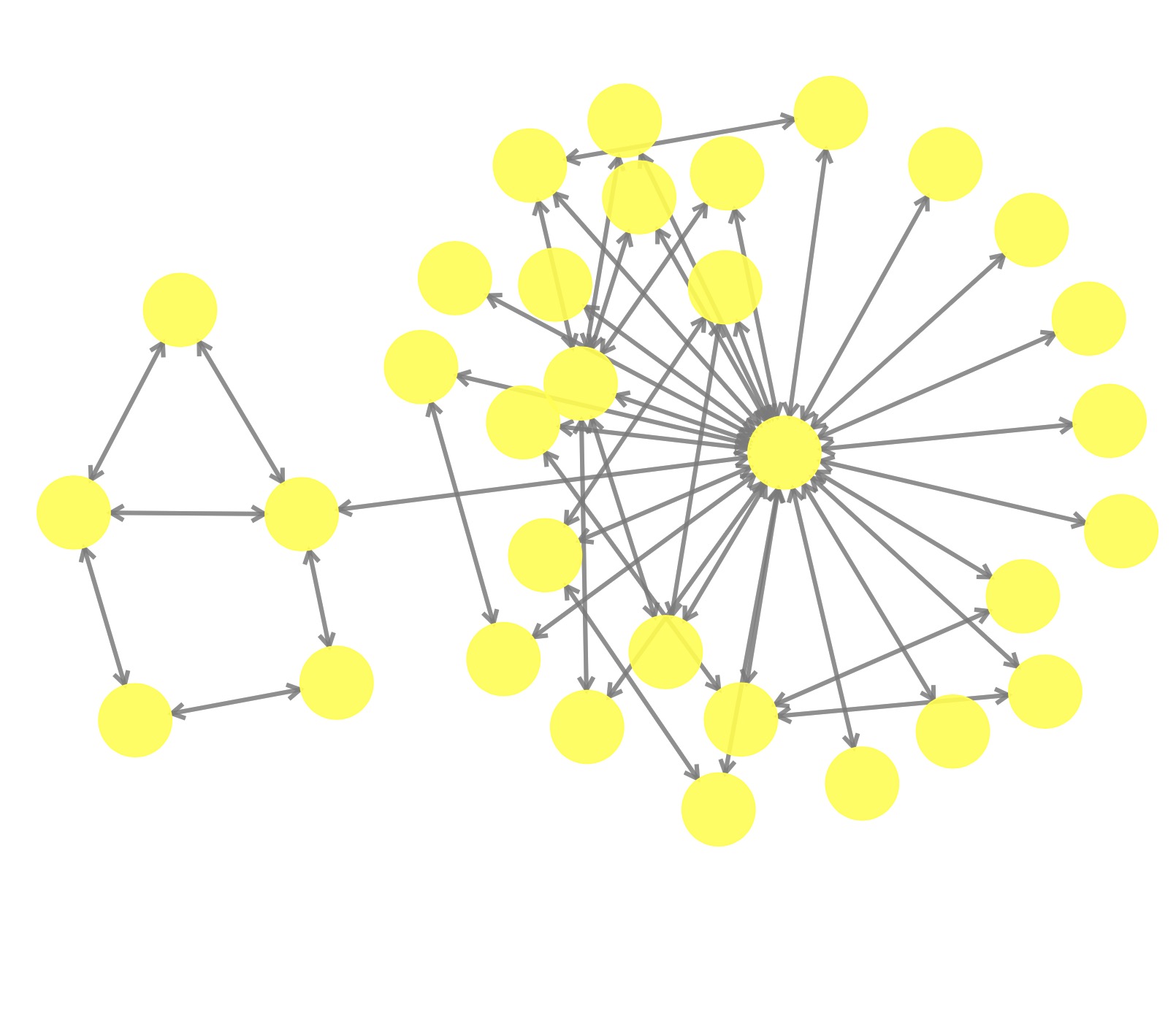} }}
    \sampleSpacing
    \subfloat[\centering BA Shapes Prediciton]{{\includegraphics[width=\sampleWidth\textwidth]{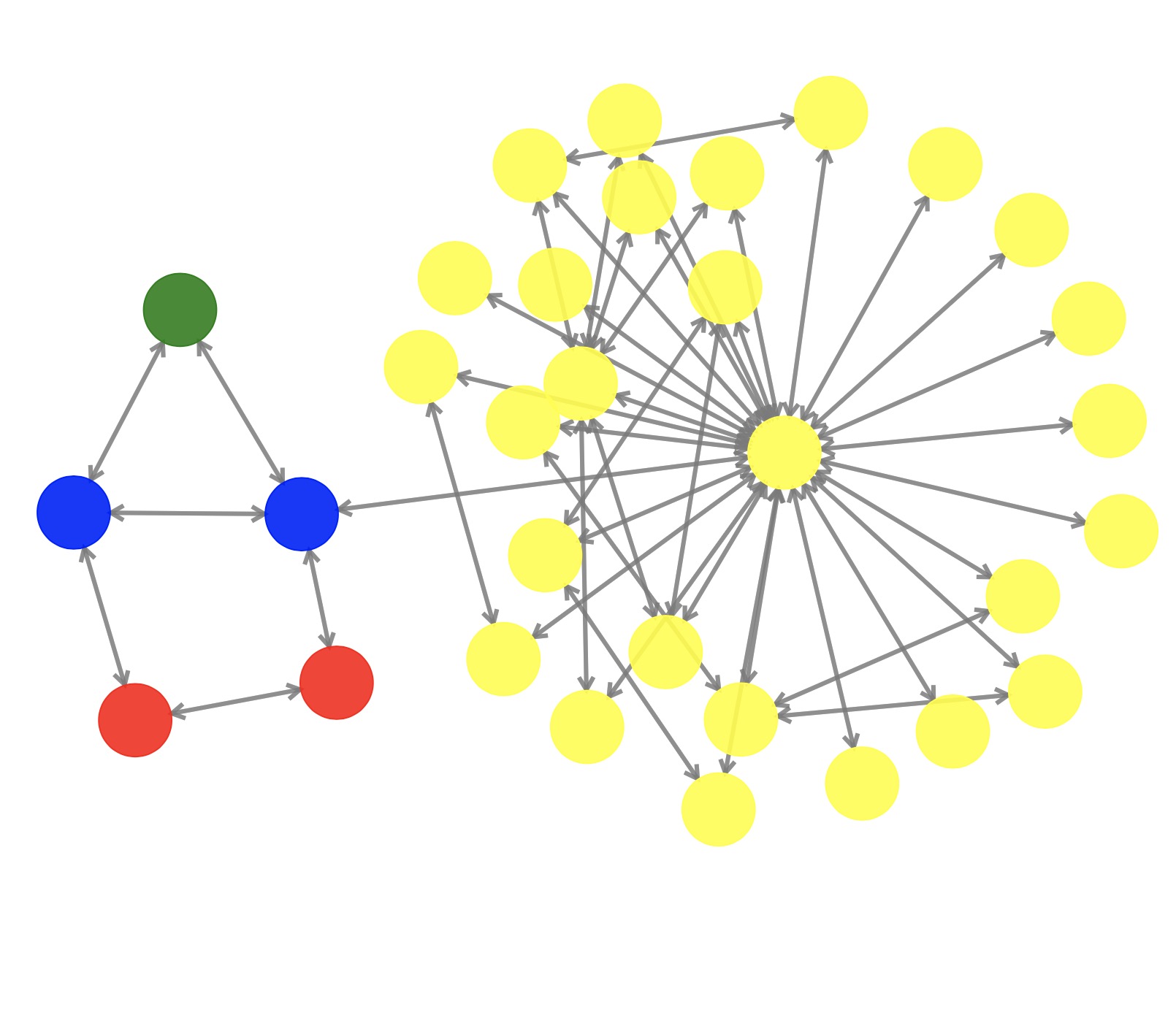} }}
    \sampleSpacing
    \subfloat[\centering Tree Grid Input]{{\includegraphics[width=\sampleWidth\textwidth]{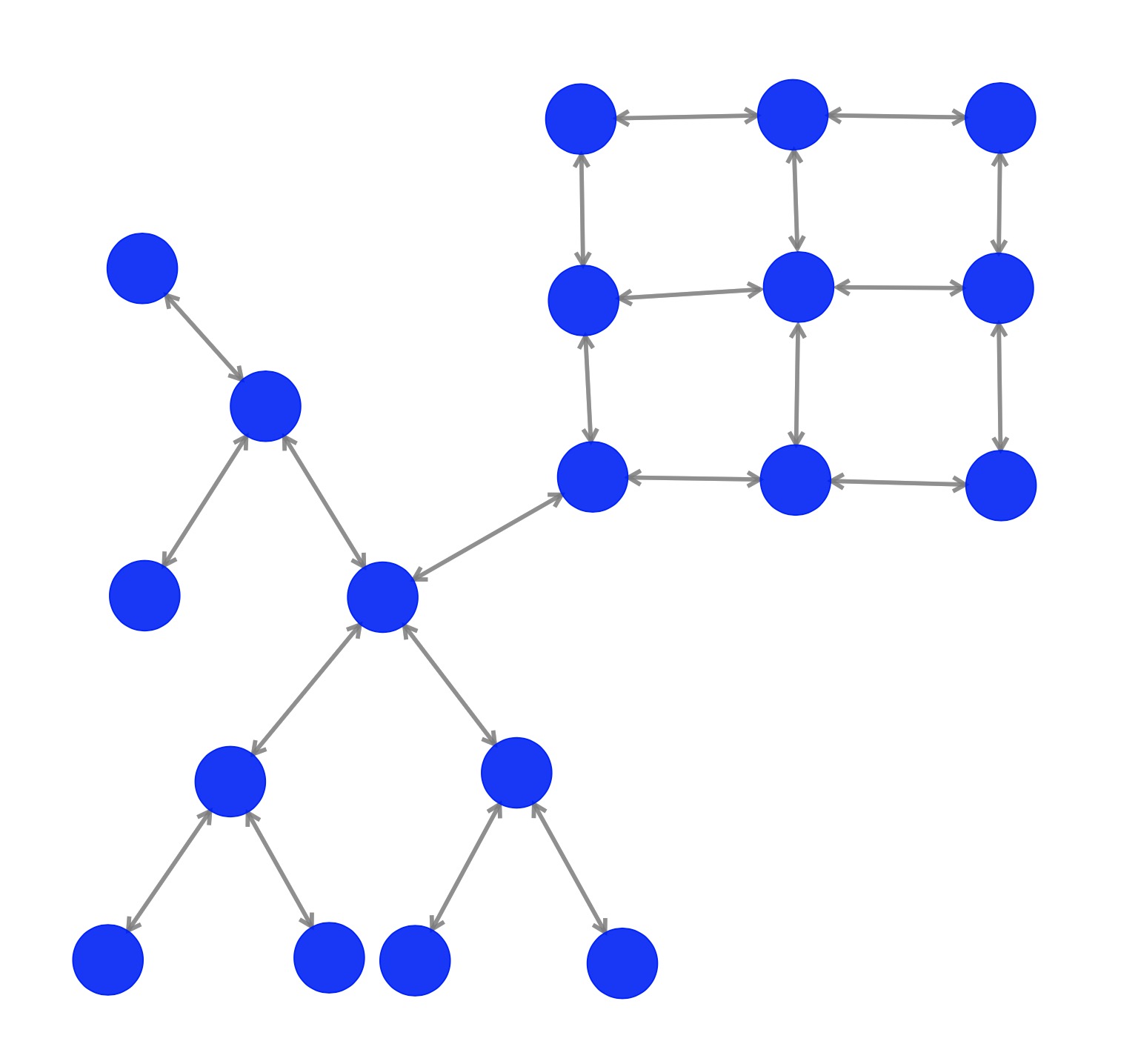}}}
    \sampleSpacing
    \subfloat[\centering Tree Grid Prediction]{{\includegraphics[width=\sampleWidth\textwidth]{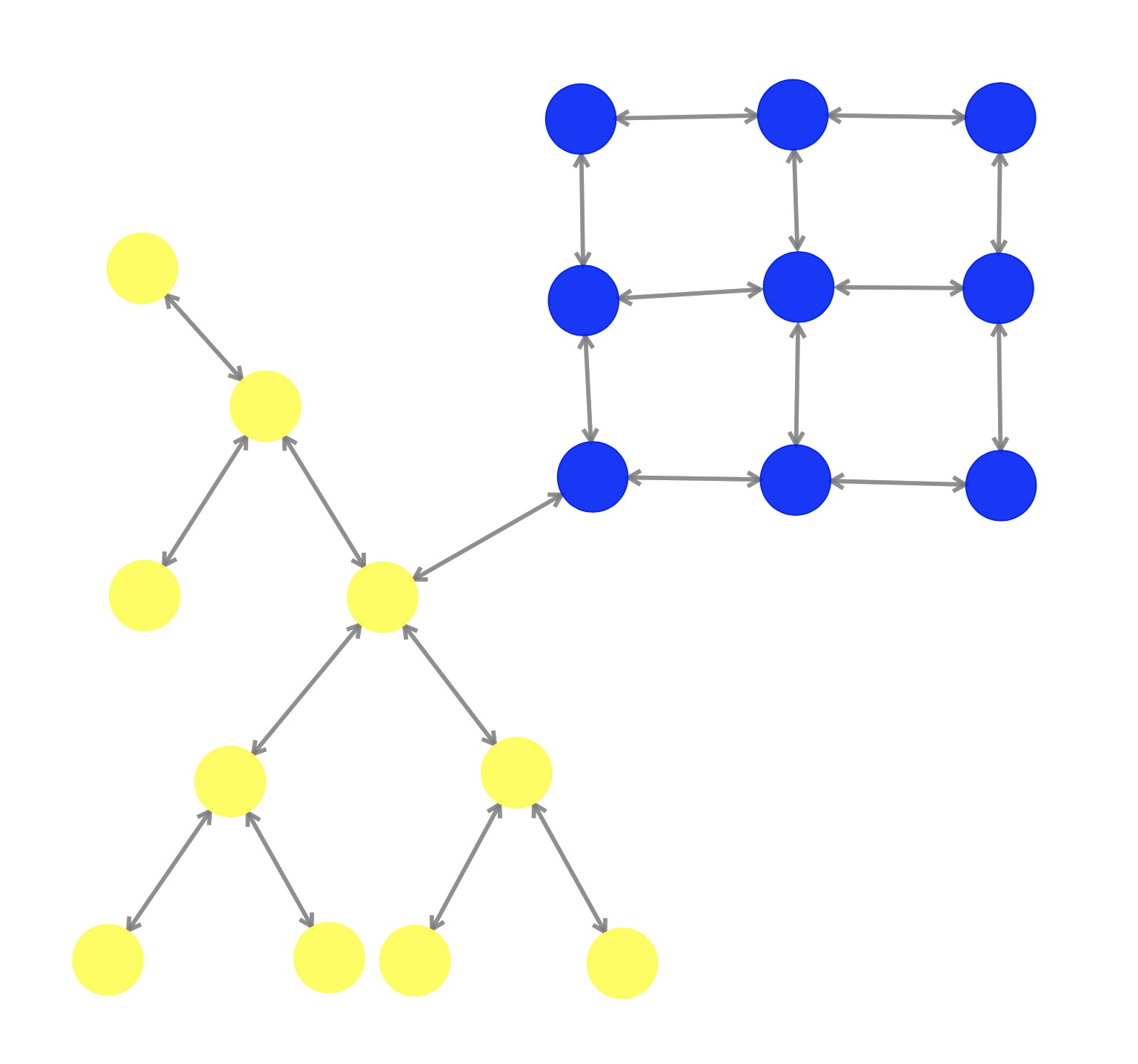}}}
    \sampleSpacing
    \subfloat[\centering BA 2Montifs - House Input]{{\includegraphics[width=\sampleWidth\textwidth]{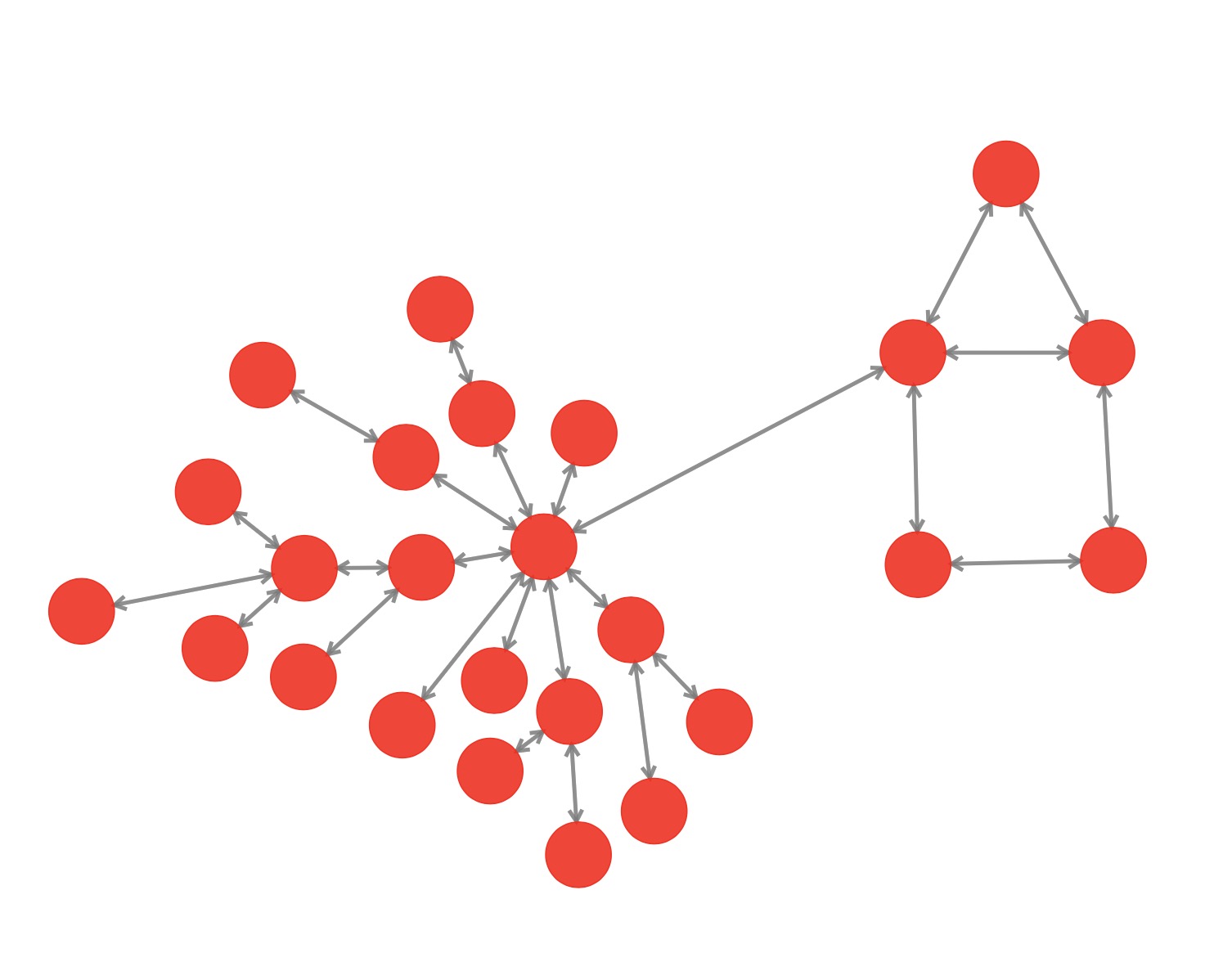}
    }}
    \sampleSpacing
    \subfloat[\centering BA 2Montifs - Cycle Input]{{\includegraphics[width=\sampleWidth\textwidth]{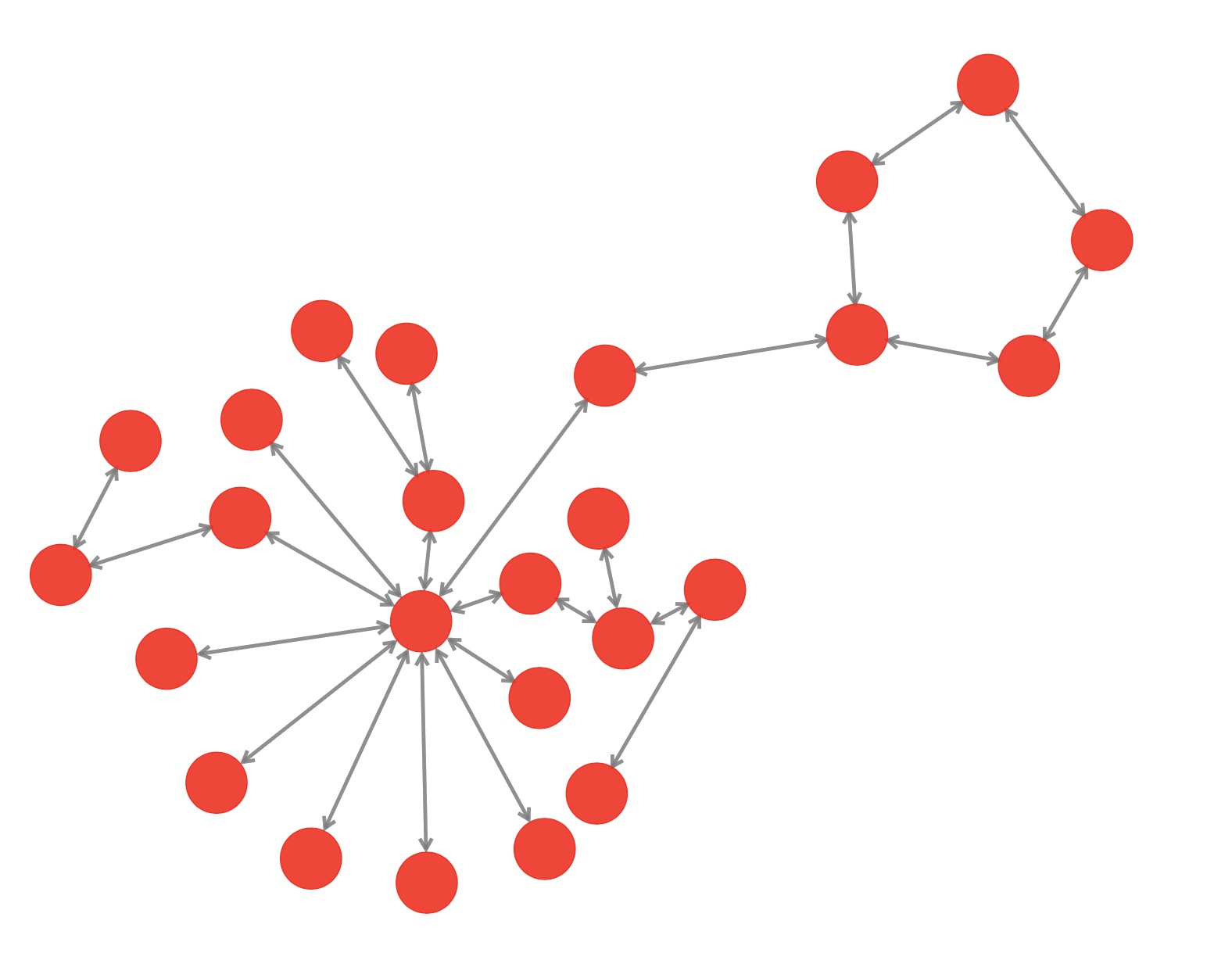}
    }}
\caption{Synthetic Benchmarks - Example Graphs} \label{fig:synthetic-examples}
\end{center}
\end{figure}

\begin{table}[!htpb]
\centering
\begin{tabular}{l r r r r r}
\hline
Dataset & Graphs & Classes & Avg. Nodes & Avg. Edges & Features \\ \hline
Infection & 1 & 7 & 1000 & 3973 & 2 \\
Negative Evidence & 1 & 2 & 2000 & 102394 & 3 \\
BA Shapes & 1 & 4 & 700 & 4110 & 0 \\
Tree Cycle & 1 & 2 & 871 & 1942 & 0 \\
Tree Grid & 1 & 2 & 1231 & 3130 & 0 \\
BA 2Motifs & 1000 & 2 & 25 & 50.96 & 0 \\
\hline
\end{tabular}
\caption{Statistics of Synthetic Datasets}
\label{table:benchmarkstats_synthetic}
\end{table}

\subsection{Real-World Datasets}

\begin{itemize}
    \item \textbf{MUTAG} \cite{debnath1991structure} is a molecule graph classification dataset. Each graph represents a nitroaromatic compound, and the goal is to predict its mutagenicity in Salmonella typhimurium. Mutagenicity is the ability of a compound to change the genetic material permanently, usually DNA, in an organism and therefore increase the frequency of mutations. The nodes in the graph represent atoms and are labeled by atom type. The edges represent bonds between atoms.
    \item \textbf{Mutagenicity} \cite{kazius2005derivation} is a molecule graph classification dataset. Each graph represents the chemical compound of a drug, and the goal is to predict its mutagenicity. The nodes in the graph represent atoms and are labeled by atom type. The edges represent bonds between atoms.
    \item \textbf{BBBP} \cite{wu2017moleculenet} is a molecule graph classification dataset. Each graph represents the chemical compound of a drug, and the goal is to predict its blood-brain barrier permeability. The nodes in the graph represent atoms and are labeled by atom type. The edges represent bonds between atoms.
    \item \textbf{PROTEINS} \cite{Borgwardt2005ProteinKernels} is a protein graph classification dataset. Each graph represents a protein that is classified as an enzyme or not and enzyme. Nodes represent the amino acids, and an edge connects two nodes if they are less than 6 Angstroms apart.
    \item \textbf{REDDIT BINARY} \cite{Borgwardt2005ProteinKernels} is a social graph classification dataset. Each graph represents the comment thread of a post on a subreddit. Nodes in the graph represent users, and there is an edge between users if one responded to at least one of the other's comments. A graph is labeled according to whether it belongs to a question/answer-based or a discussion-based subreddit.
    \item \textbf{IMDB BINARY} \cite{Borgwardt2005ProteinKernels} is a social graph classification dataset. Each graph represents the ego network of an actor/actress. In each graph, nodes represent actors/actresses, and there is an edge between them if they appear in the same film. A graph is labeled according to whether the actor/actress belongs to the Action or Romance genre.
    \item \textbf{COLLAB} \cite{Borgwardt2005ProteinKernels} is a social graph classification dataset. A graph represents a researcher's ego network. The researcher and their collaborators are nodes, and an edge indicates collaboration between two researchers. A graph is labeled according to whether the researcher belongs to the field of high-energy physics, condensed matter physics, or astrophysics.
    
\end{itemize}

\begin{figure}[htb]
\begin{center}
\centering
    \subfloat[\centering MUTAG Input]{{\includegraphics[width=\sampleWidth\textwidth]{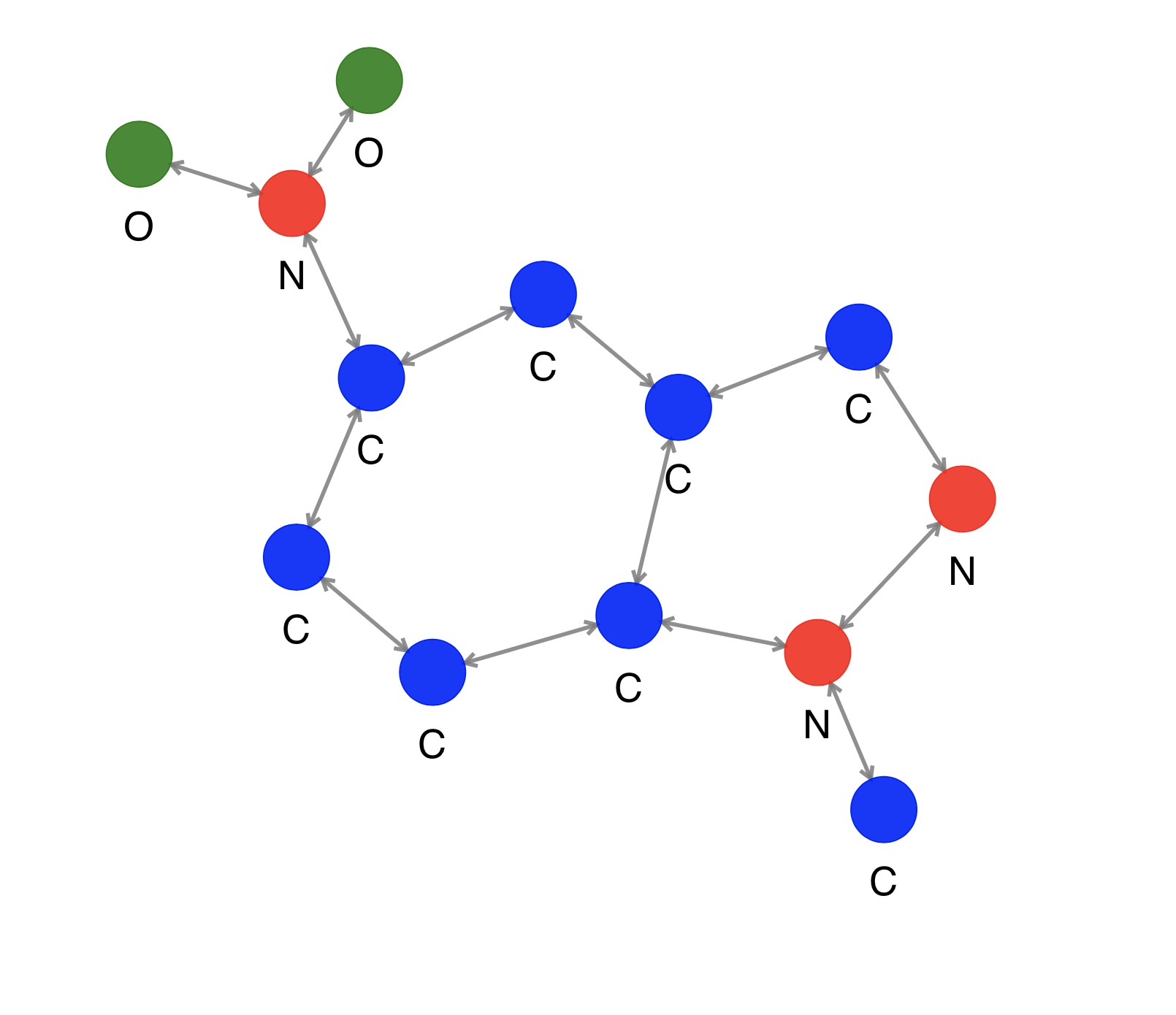} }}
    \sampleSpacing
    \subfloat[\centering Mutagenicity Input]{{\includegraphics[width=\sampleWidth\textwidth]{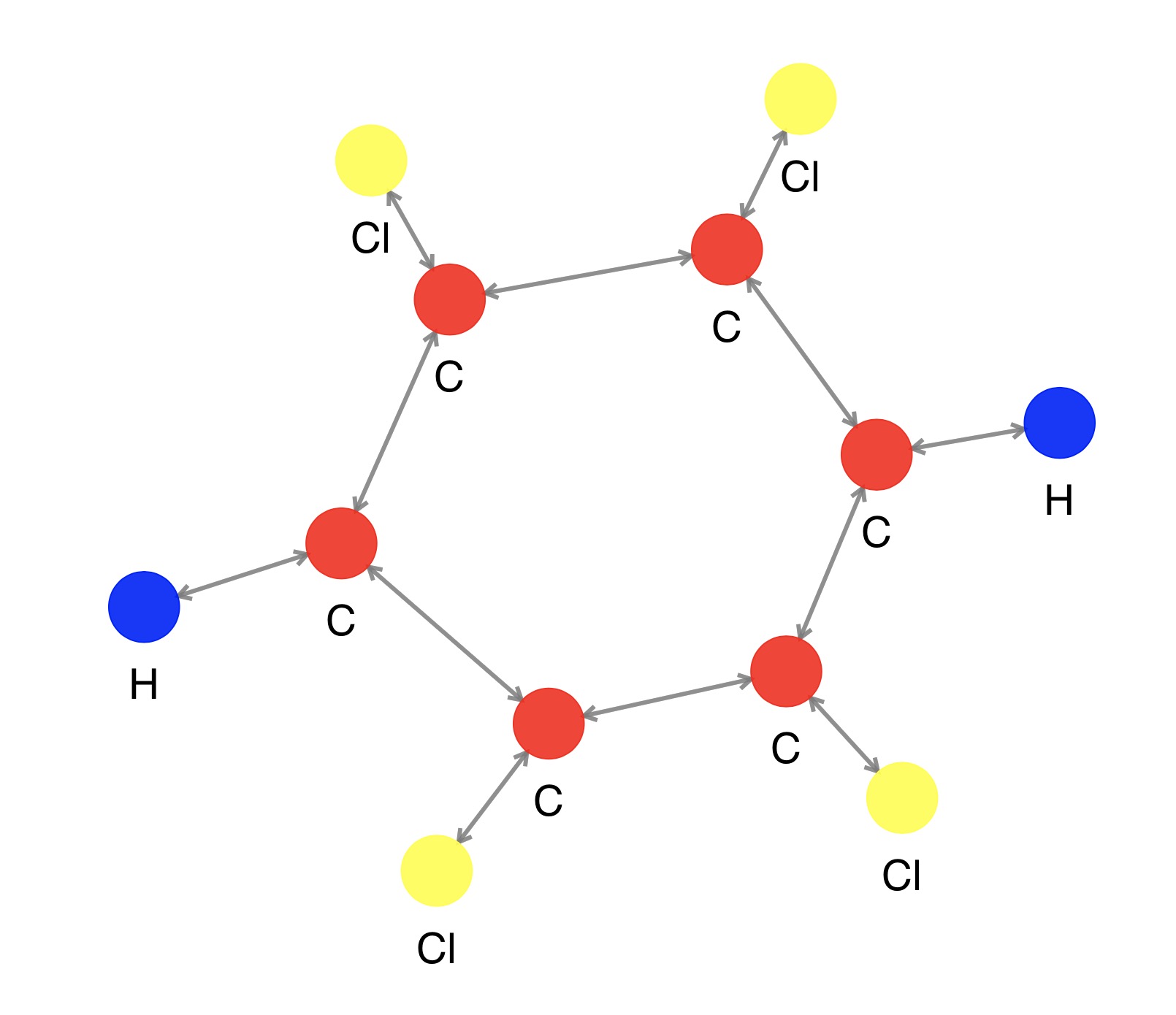} }}
    \sampleSpacing
    \subfloat[\centering PROTEINS Input]{{\includegraphics[width=\sampleWidth\textwidth]{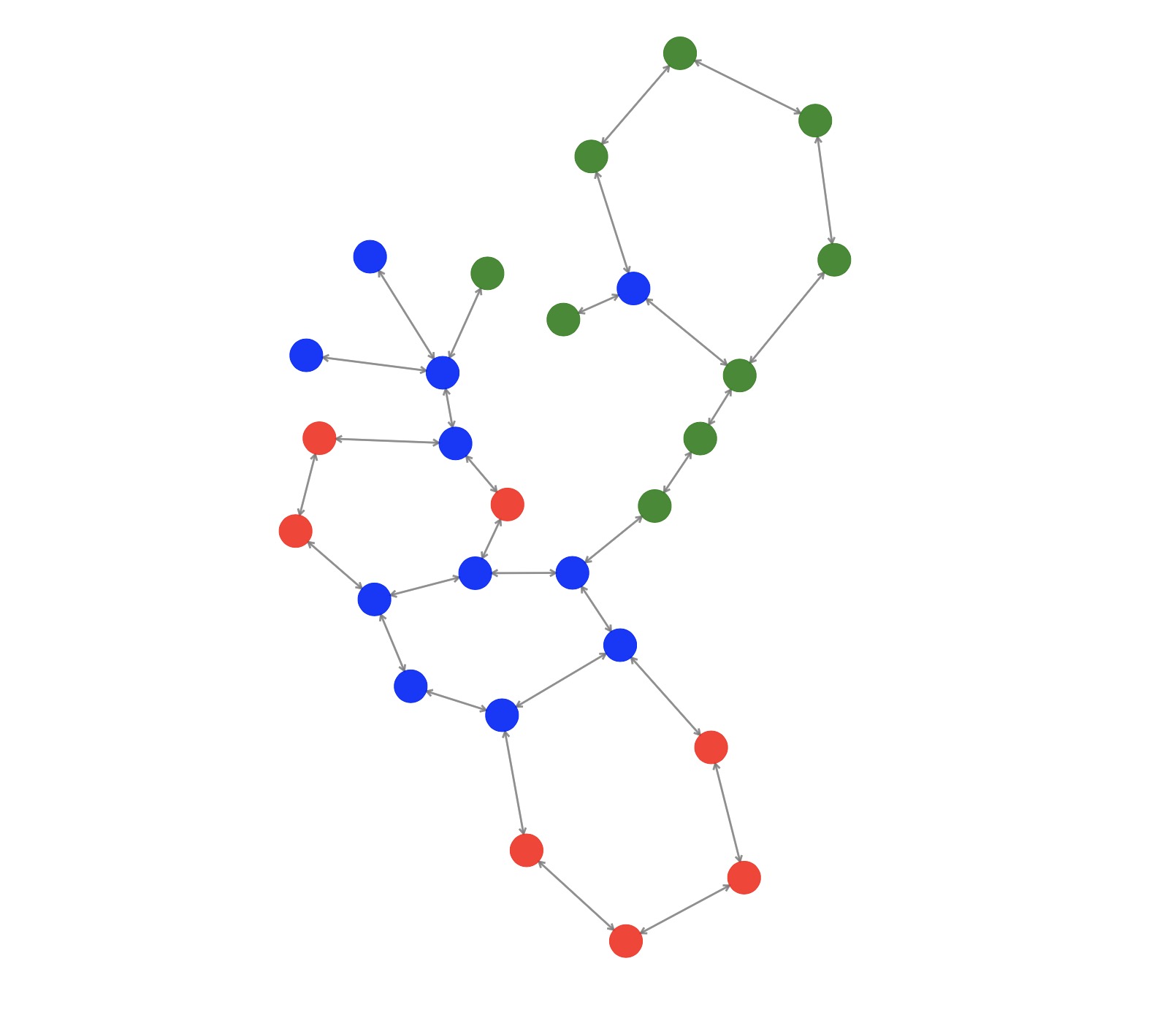}}}
    \\%\sampleSpacing
    \subfloat[\centering Reddit Binary Discussion]{{\includegraphics[width=\sampleWidth\textwidth]{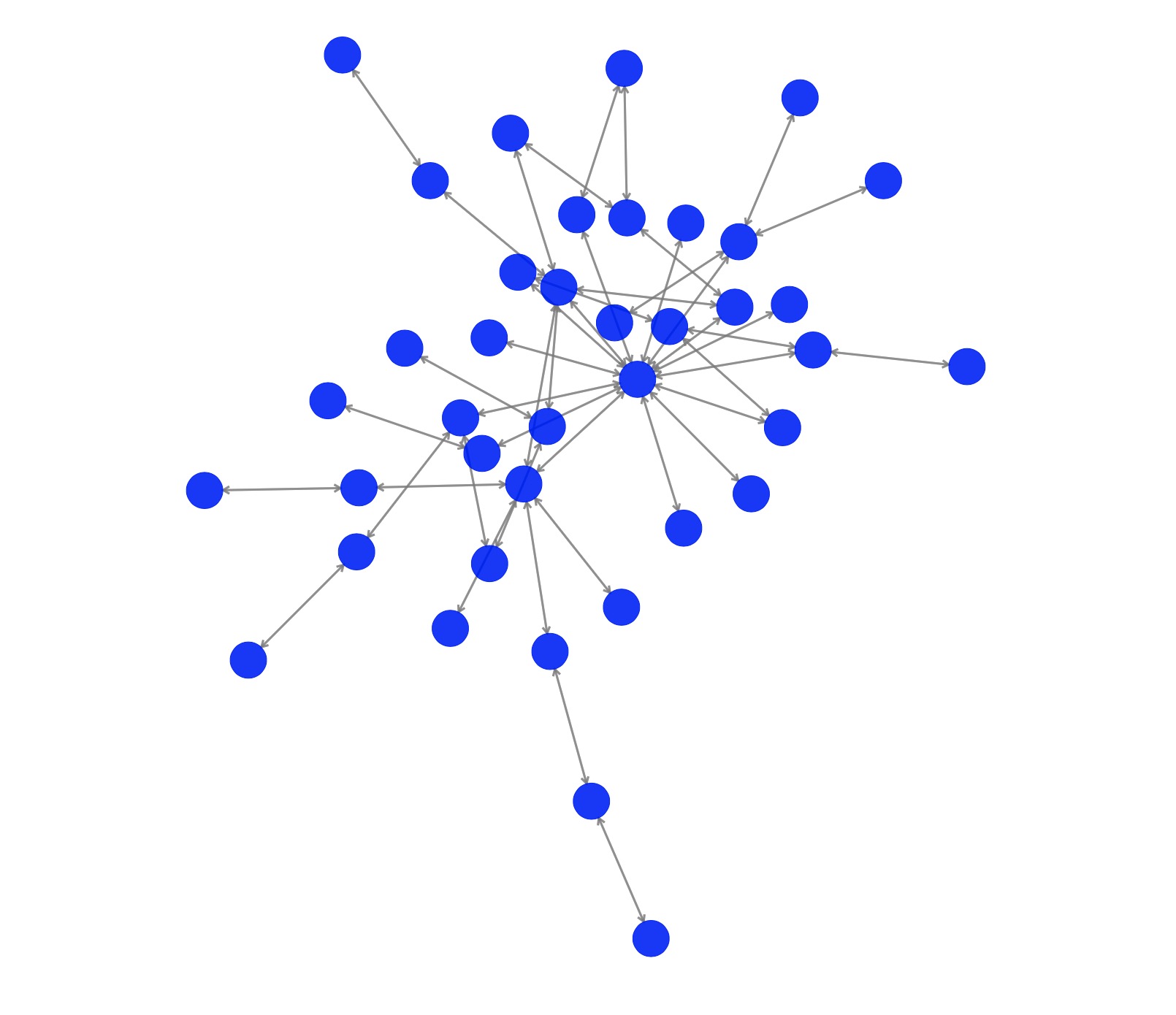} }}
    \sampleSpacing
    \subfloat[\centering Reddit Binary Q/A]{{\includegraphics[width=\sampleWidth\textwidth]{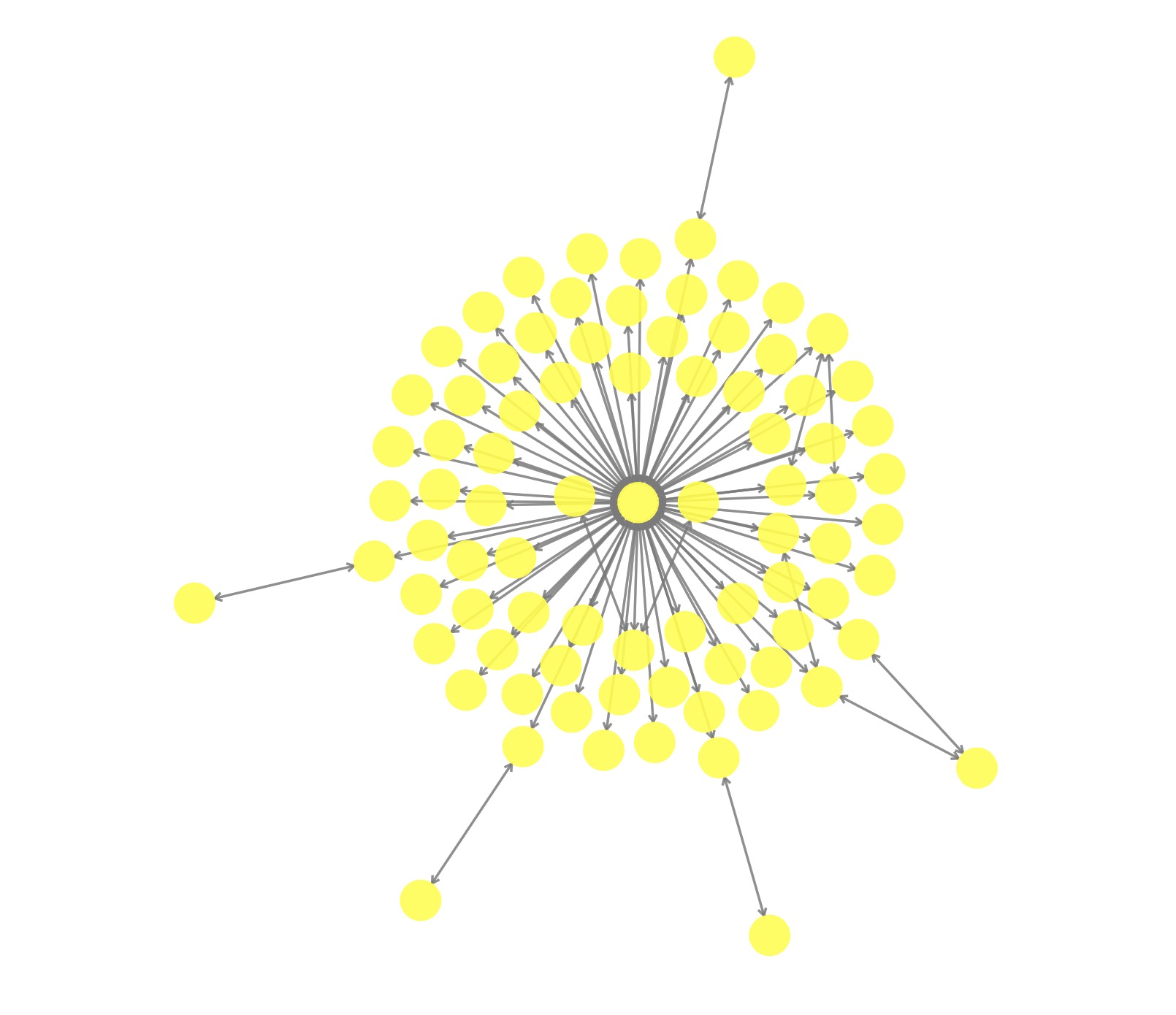} }}
    \sampleSpacing
    \subfloat[\centering IMDB Binary Input]{{\includegraphics[width=\sampleWidth\textwidth]{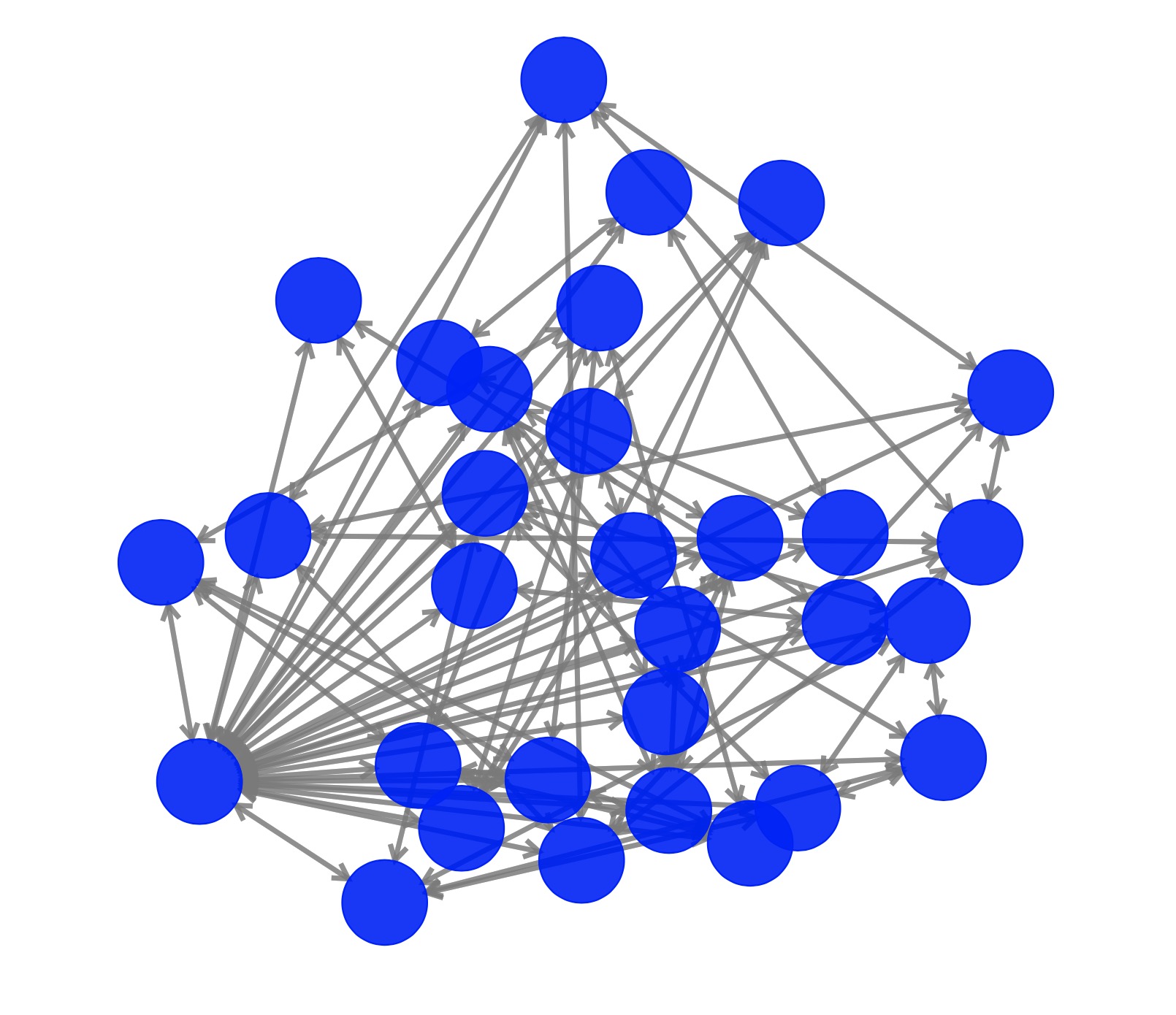} }}
    \sampleSpacing
    \subfloat[\centering COLLAB Input]{{\includegraphics[width=\sampleWidth\textwidth]{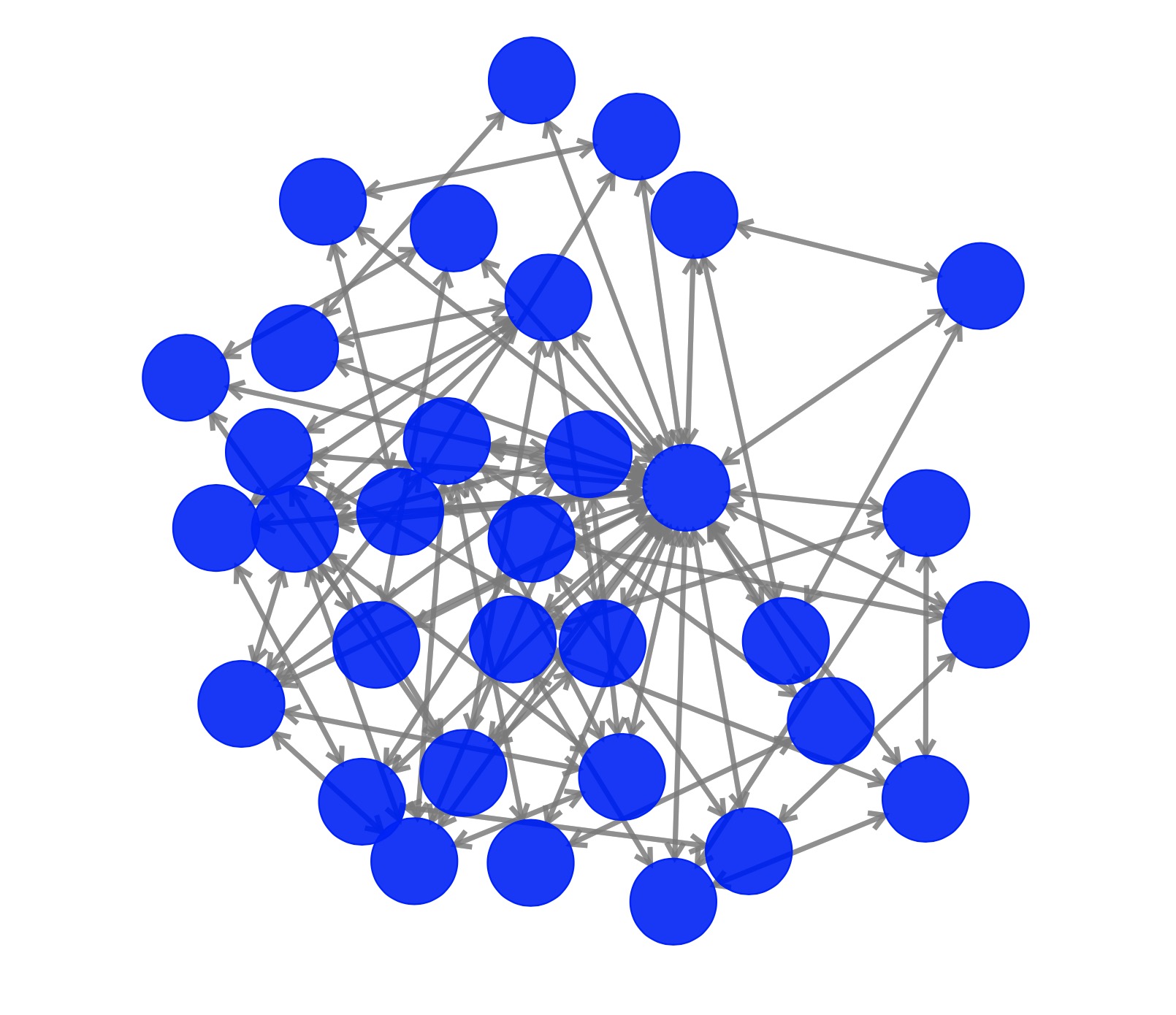} }}
\caption{Real-world benchmarks - Example graphs} \label{fig:real-world-examples}
\end{center}
\end{figure}

\begin{table}[!htpb]
\centering
\begin{tabular}{l r r r r r}
\hline
Dataset & Graphs & Classes & Avg. Nodes & Avg. Edges & Features \\ \hline
MUTAG & 188 & 2 & 17.93 & 39.59 & 7 \\
Mutagenicity & 4337 & 2 & 30.32 & 61.54 & 14 \\
BBBP & 2039 & 2 & 24.06 & 51.91 & 9 \\
PROTEINS & 1113 & 2 & 39.06 & 145.63 & 3 \\
REDDIT BINARY & 2000 & 2 & 429.63 & 995.51 & 0 \\
IMDB BINARY & 1000 & 2 & 19.77 & 193.06 & 0 \\
COLLAB & 5000 & 3 & 74.49 & 4914.43 & 0 \\
\hline
\end{tabular}
\caption{Statistics of Real-World Datasets}
\label{table:benchmarkstats_realworld}
\end{table}

\clearpage
\section{More Qualitative Experiments}
\label{sec:more_example_interpretations}

\subsection{BA-Shapes}
\begin{figure}[h!]
    \begin{subfigure}[t]{0.23\textwidth}
        \centering
        \includegraphics[width=\textwidth]{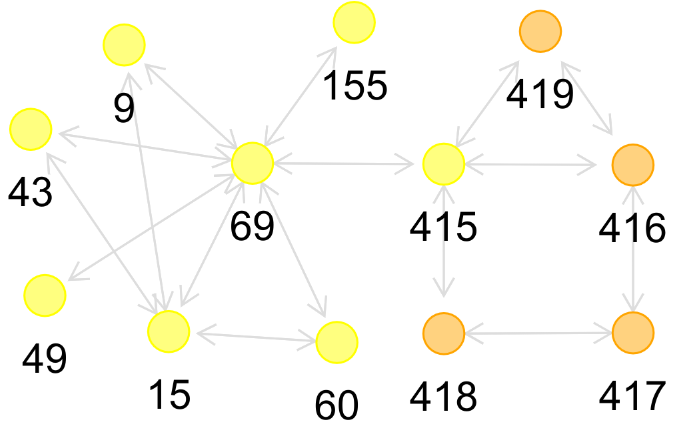}
        \caption{}
        \label{fig:bashapes1}
    \end{subfigure}\hfill%
    \begin{subfigure}[t]{0.25\textwidth}
        \centering
        \includegraphics[width=\textwidth]{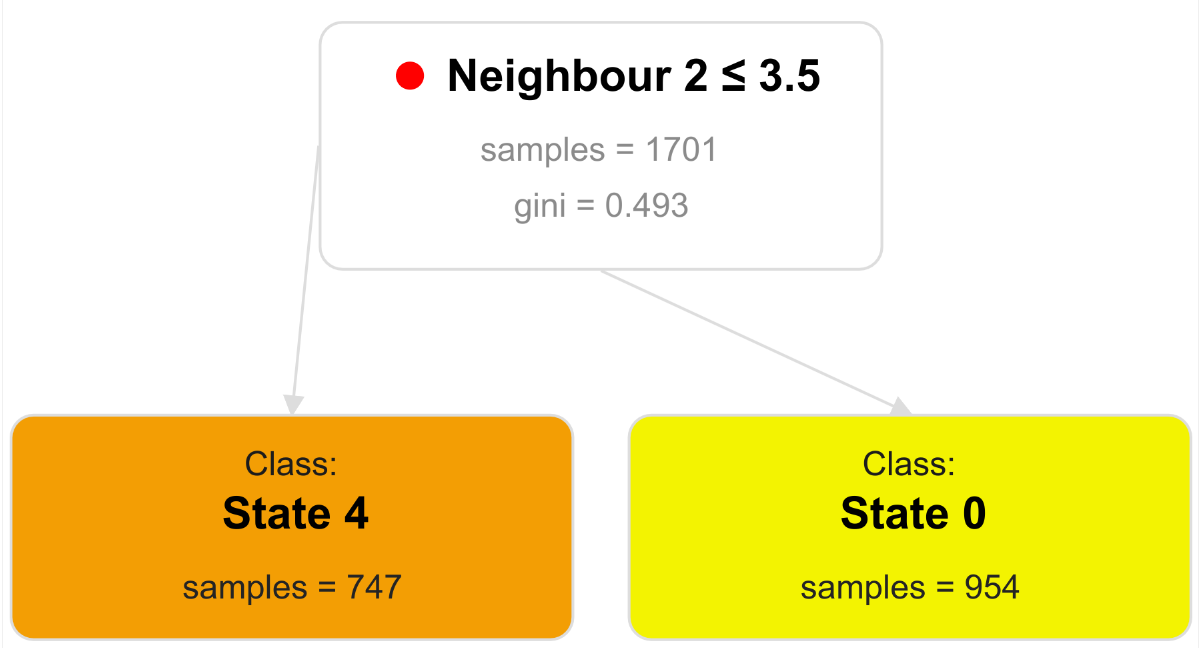}
        \caption{}
        \label{fig:bashapestree1}
    \end{subfigure}\hfill%
    \begin{subfigure}[t]{0.23\textwidth}
        \centering
        \includegraphics[width=\textwidth]{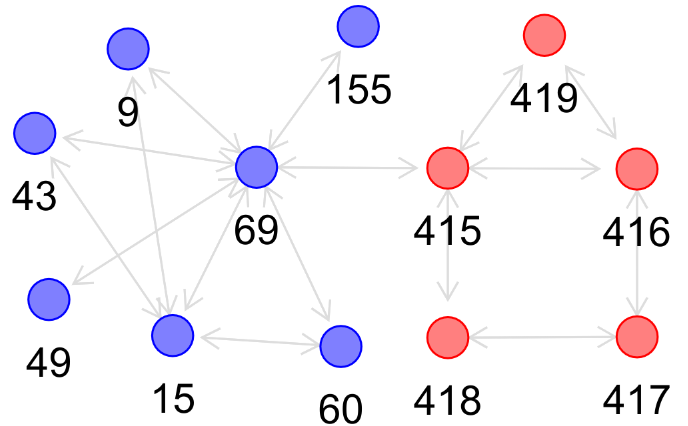}
        \caption{}
        \label{fig:bashapes2}
    \end{subfigure}\hfill%
    \begin{subfigure}[t]{0.25\textwidth}
        \centering
        \includegraphics[width=\textwidth]{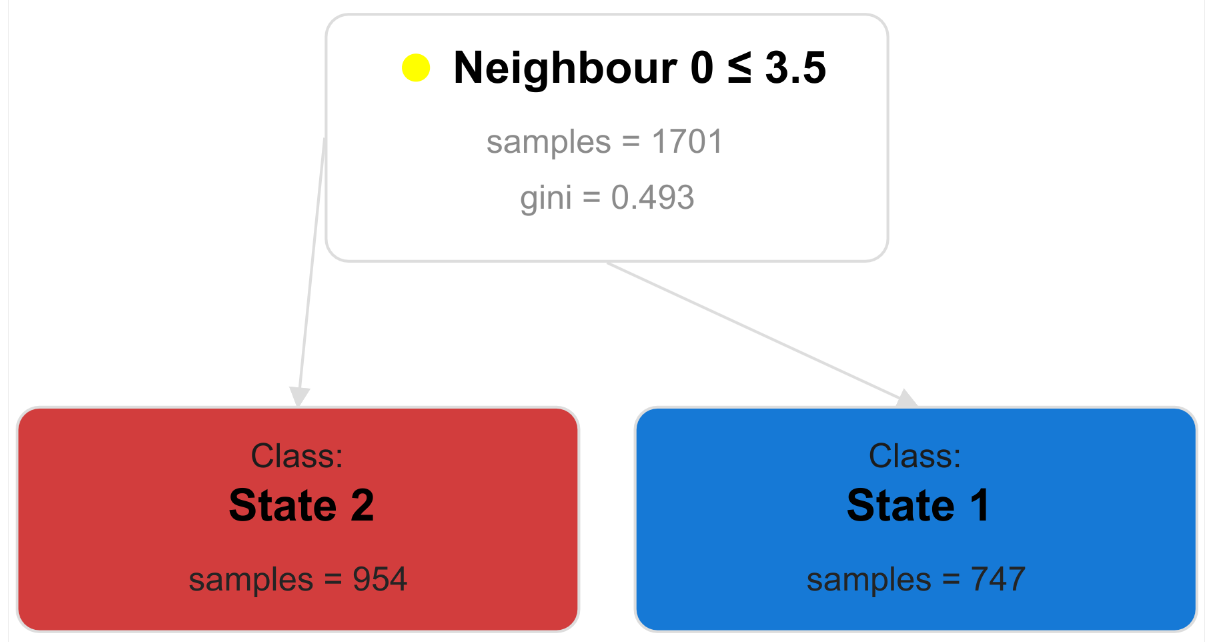}
        \caption{}
        \label{fig:bashapestree2}
    \end{subfigure}\\\hfill% 
    \hspace*{0.12\textwidth}
    \begin{subfigure}[t]{0.23\textwidth}
        \centering
        \includegraphics[width=\textwidth]{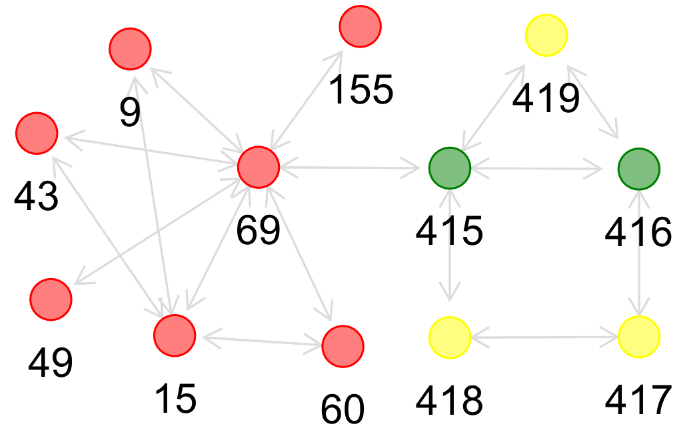}
        \caption{}
        \label{fig:bashapes3}
    \end{subfigure}\hfill%
    \begin{subfigure}[t]{0.33\textwidth}
        \centering
        \includegraphics[width=\textwidth]{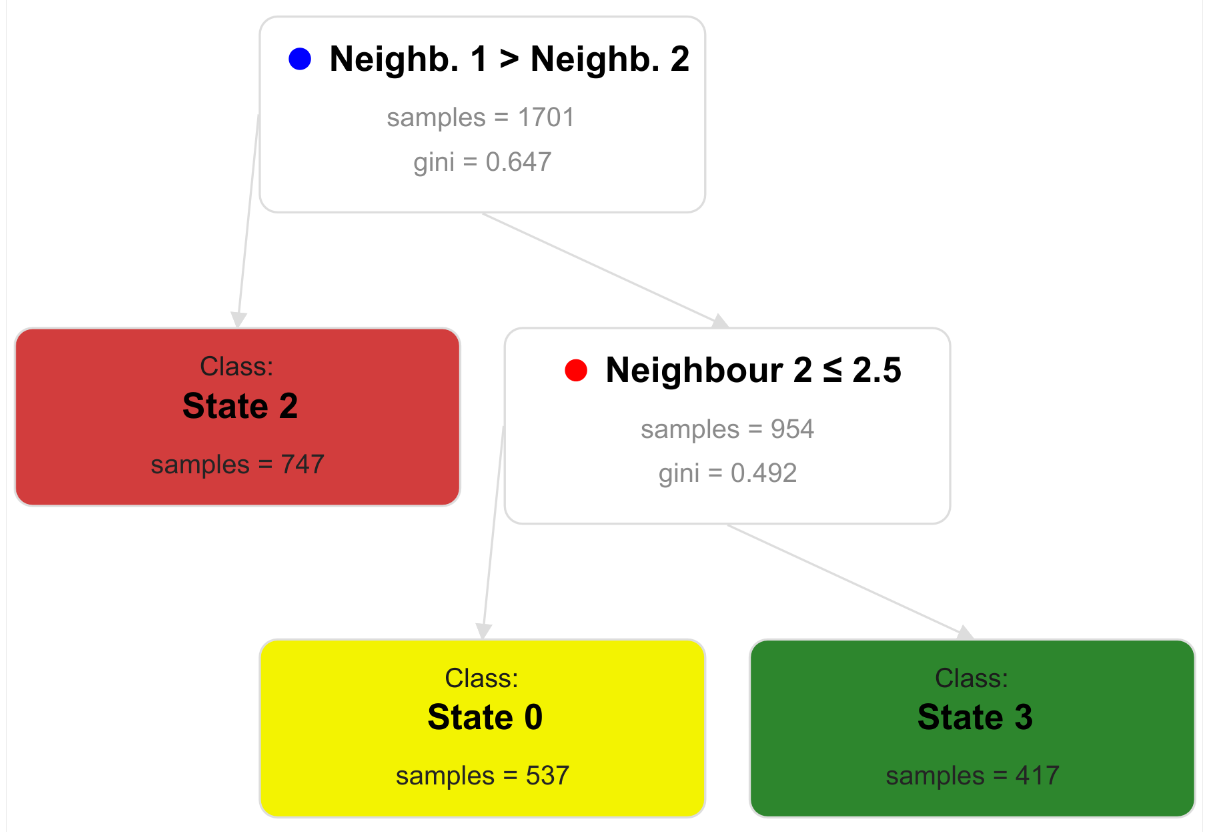}
        \caption{}
        \label{fig:bashapestree3}
    \end{subfigure}\hspace*{0.12\textwidth}\\
    \hspace*{0.12\textwidth}
    \begin{subfigure}[t]{0.23\textwidth}
        \centering
        \includegraphics[width=\textwidth]{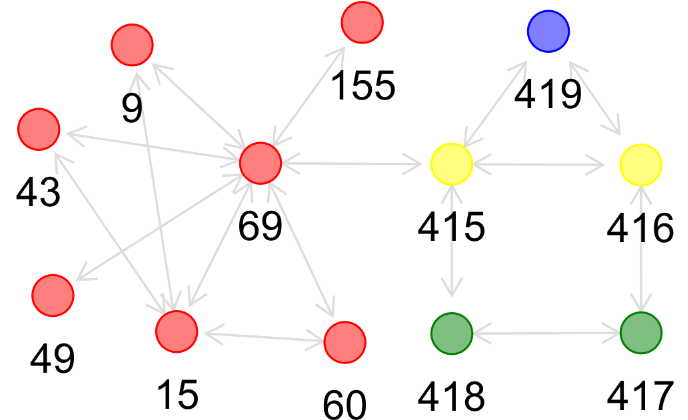}
        \caption{}
        \label{fig:bashapes4}
    \end{subfigure}
    \hfill%
    \begin{subfigure}[t]{0.4\textwidth}
        \centering
        \includegraphics[width=\textwidth]{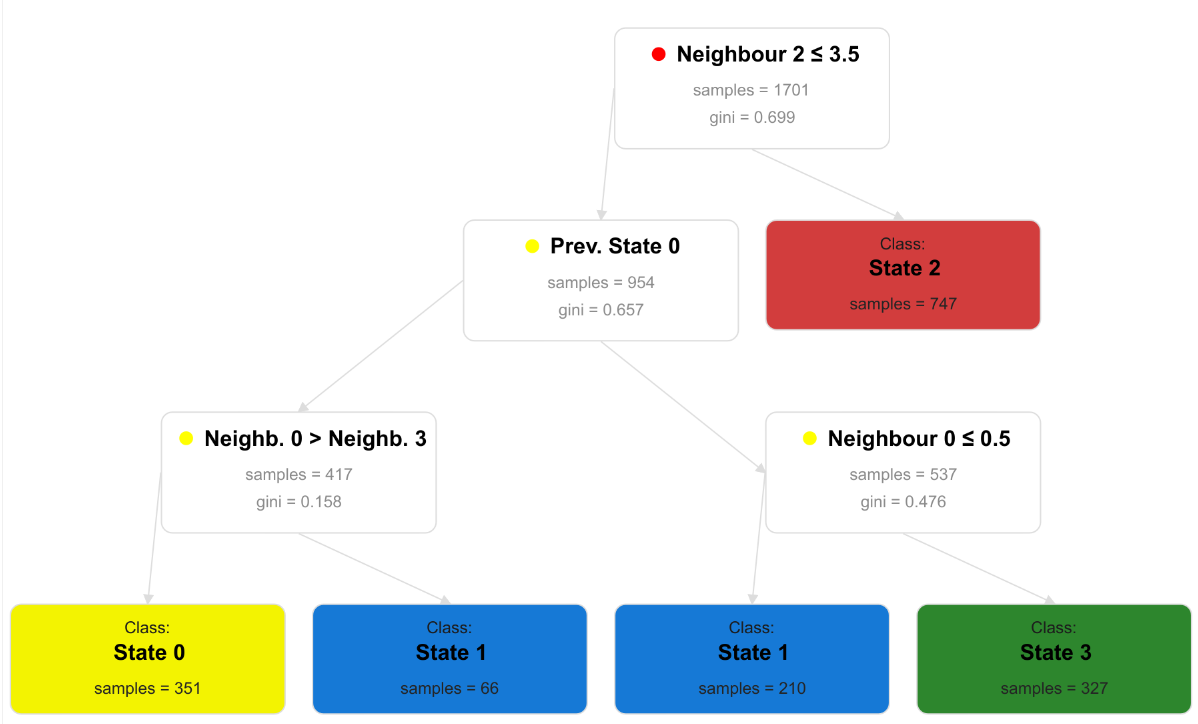}
        \caption{}
        \label{fig:bashapestree4}
    \end{subfigure}\hspace*{0.09\textwidth}%
    \caption{DT+GNN solves an instance of the BA-shapes dataset. Figures (a), (c), (e), and (g) show the different states of nodes through different layers, and the other plots (b), (d), (f), and (h) show the respective decision trees. Numbers denote node IDs. (a)-(d) First, DT+GNN gets an idea of where the house is by finding low-degree nodes. (e) and (f) Inside the house, DT+GNN finds the nodes with a degree of $2$; the nodes with a higher degree form the center. (g) and (h) The lonely node is the top of the house while the two connected degree $2$ nodes are the bottom.}
\end{figure}

We first want to discuss the approach learned by DT+GNNs to solving the BA-Shapes dataset~\citep{ying2019gnnexplainer}. The goal is to identify the nodes in a house (bottom, middle, top) versus the underlying base graph. The base graph is more densely connected than the house so DT+GNN first identifies nodes with at most degree $3$ (Figure~\ref{fig:bashapes1}). This identifies almost every node in the house but the ones that connect the house to the graph. These nodes are found in the next step~(\ref{fig:bashapes2}). From here, DT+GNN needs to map the nodes in the house to their respective positions. It starts by finding the two middle nodes with a degree larger than $2$~(Figure~\ref{fig:bashapes3}). Last, it can distinguish the top from the bottom nodes, since the top has no other degree $2$ neighbor~(Figure~\ref{fig:bashapes4}).
\label{fig:bashapes}

\subsection{Tree-Grid}
The Tree-Grid~\citep{ying2019gnnexplainer} dataset is similar to the Tree-Cycles dataset we discussed in the main body of the paper. The base graph is a balanced binary tree to which we append $3\times 3$ grids. As in the Tree-Cycles example, a GNN does not need to see the whole grid to make a prediction.

\begin{figure}[h!]
    \begin{subfigure}{0.23\textwidth}
        \centering
        \includegraphics[width=\textwidth]{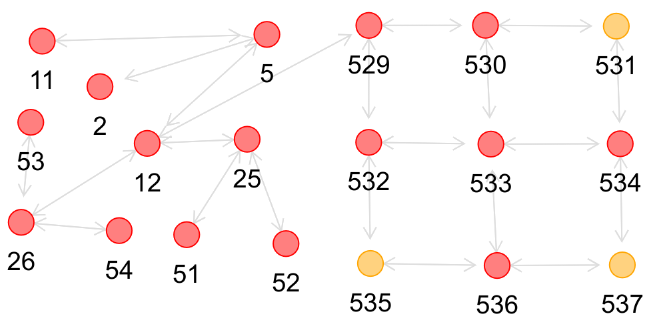}
        \caption{}
        \label{fig:grid1}
    \end{subfigure}\hfill%
    \begin{subfigure}{0.23\textwidth}
        \centering
        \includegraphics[width=\textwidth]{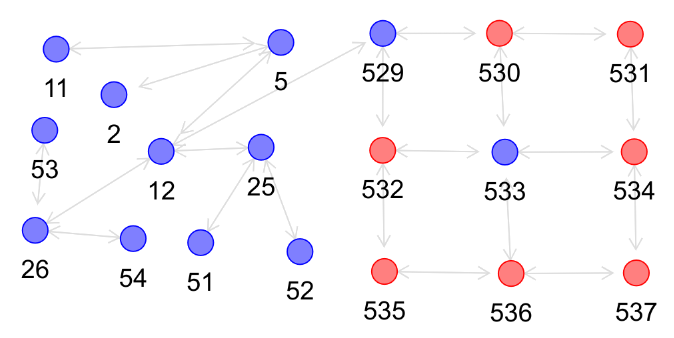}
        \caption{}
        \label{fig:grid2}
    \end{subfigure}\hfill%
    \begin{subfigure}{0.23\textwidth}
        \centering
        \includegraphics[width=\textwidth]{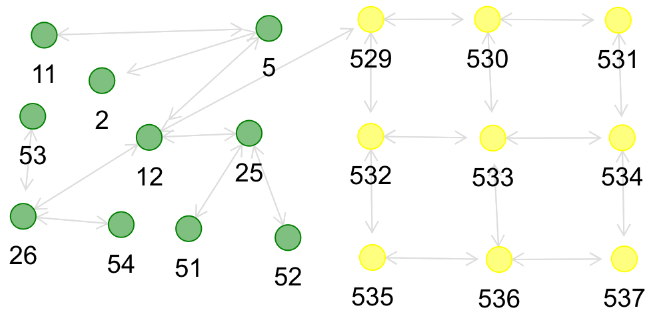}
        \caption{}
        \label{fig:grid3}
    \end{subfigure}\hfill%
    \begin{subfigure}{0.23\textwidth}
        \centering
        \includegraphics[width=\textwidth]{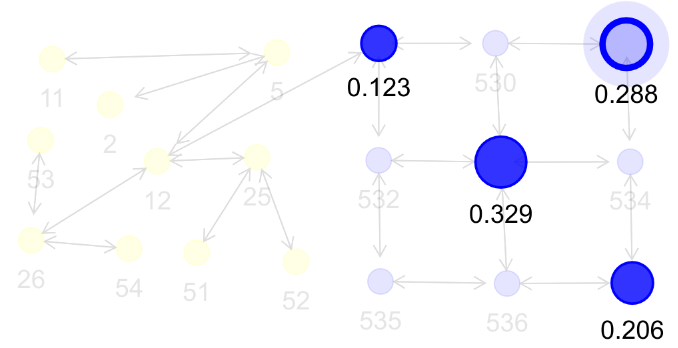}
        \caption{}
        \label{fig:gridexpl}
    \end{subfigure}\hfill%
    \caption{DT+GNN solves an instance of the Tree-Grid dataset. Figures (a)-(c) show the progression of node states for the first three layers. Numbers denote node IDs. (a) DT+GNN finds nodes of degree $2$ in the first layer, the corners of the grid. (b) DT+GNN finds the neighbors of the corner nodes. (c) DT+GNN finds the nodes next to the previous nodes --- completing the grid. (d) This plot shows the explanation scores for the top-right node. Explanations are the $2-$distance neighborhood. Importantly, the bottom-left node in the grid was never seen by the top-right node --- rightfully it is not part of the explanation.}
\end{figure}

Because the base graph is a balanced binary tree, only the root node and corner nodes in the grid have a degree of $2$. DT+GNN first finds the corner nodes (Figure~\ref{fig:grid1}). From there, DT+GNN can incrementally build the grid by adding the neighboring nodes, finishing after two sets of expansion (Figures~\ref{fig:grid2} and \ref{fig:grid3}). Similar to the Tree-Cycle example, we do not need to read the whole grid structure. Therefore, it would be wrong to expect an explanation method to highlight all of the grid. For example, DT+GNN's explanation does not use the node in the $4-$hop neighborhood in the opposite corner in the explanation (Figure~\ref{fig:gridexpl}) since it did not look that far.
\label{fig:treegrid_example}

\subsection{MUTAG}
\begin{figure}[h!]
    \begin{subfigure}{0.35\textwidth}
        \centering
        \includegraphics[width=\textwidth]{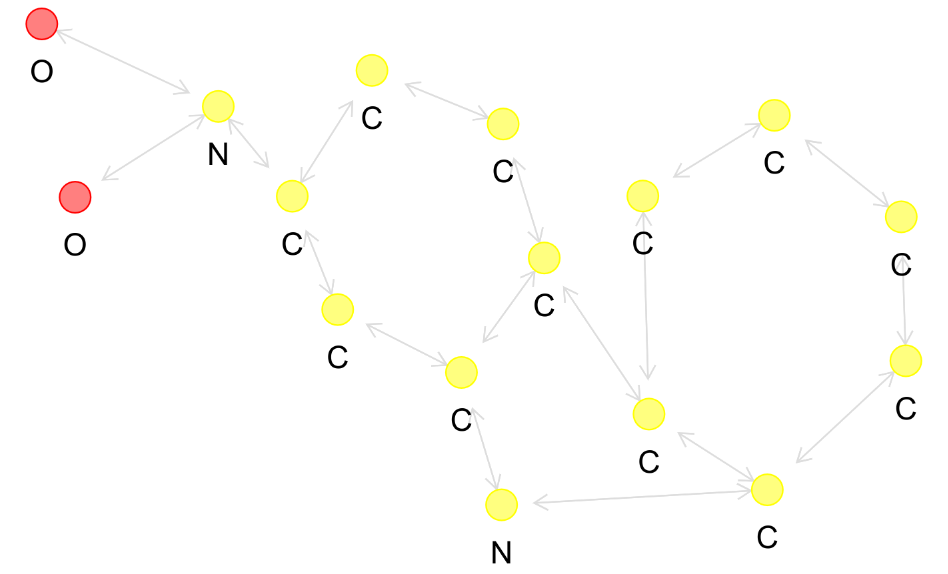}
        \caption{}
        \label{fig:mutag1}
    \end{subfigure}\hfill%
    \begin{subfigure}{0.35\textwidth}
        \centering
        \includegraphics[width=\textwidth]{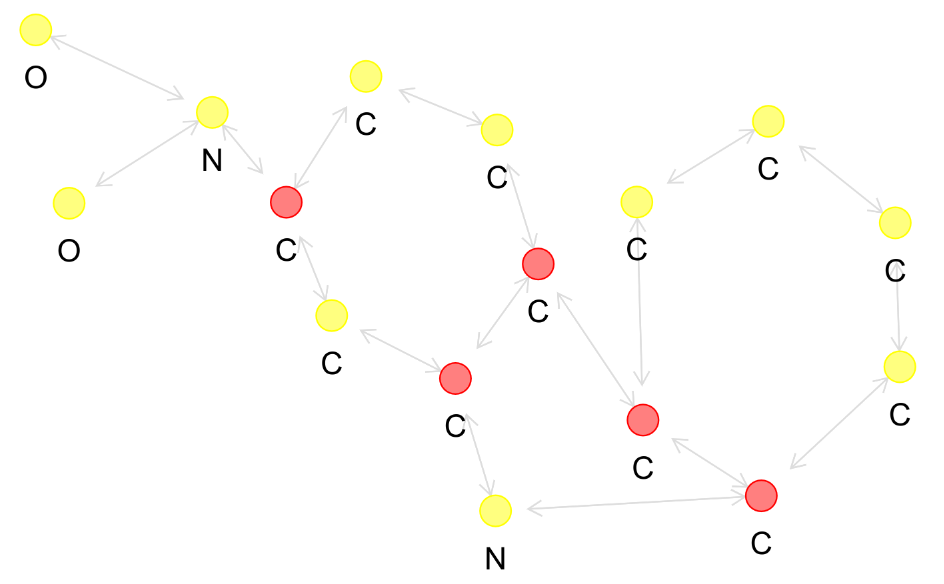}
        \caption{}
        \label{fig:mutag2}
    \end{subfigure}\hfill%
    \begin{subfigure}{0.23\textwidth}
        \centering
        \includegraphics[width=\textwidth]{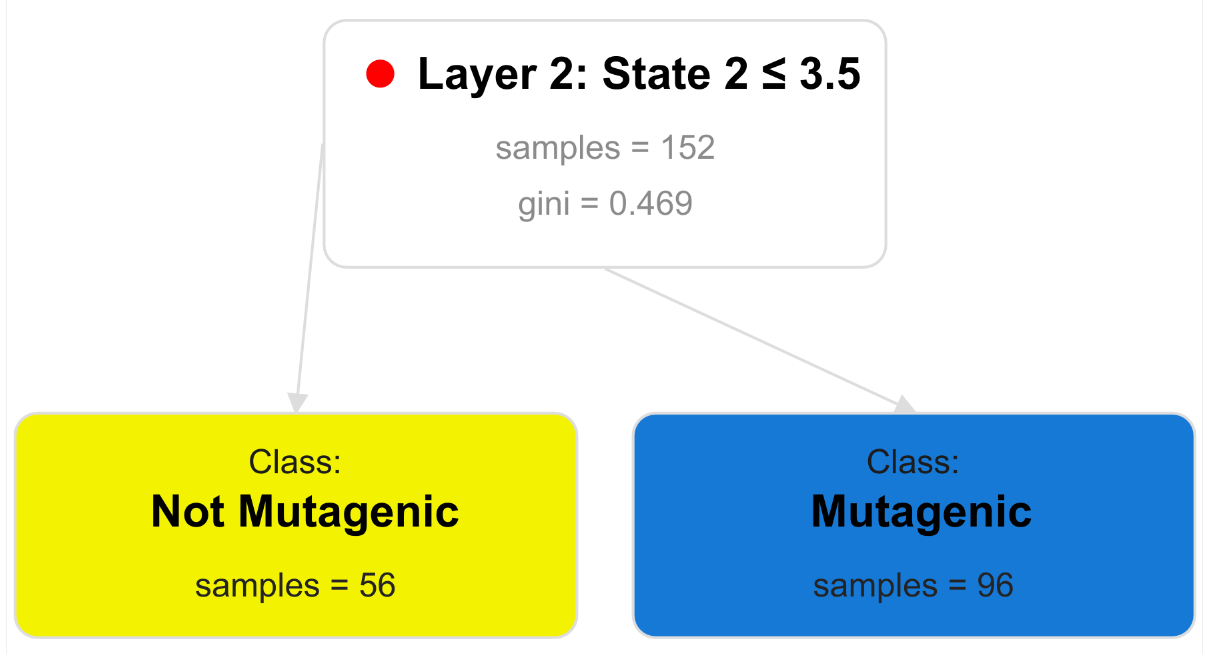}
        \caption{}
        \label{fig:mutagtree}
    \end{subfigure}\hfill%
    \caption{DT+GNN solves an instance of the MUTAG dataset. (a) and (b) show the learned node states for the first two layers. Numbers denote node IDs. (a) DT+GNN finds $O$ atoms by a degree count (b) DT+GNN finds nodes that have $3$ connections but not to $O$ atoms. (c) shows the final decoder layer. If there are $4$ or red nodes from the second layer (see (b)), the graph is mutagenic.}
\end{figure}

As discussed in the main body, the presence of $NO_2$ groups in MUTAG~\citep{debnath1991structure} is not informative for mutagenicity since all graphs have such a group. Instead, DT+GNN counts the nodes that have $3$ neighbors of a degree of $2$ or more (Figure~\ref{fig:mutag2}). Therefore, DT+GNN is looking for atoms that are not connected to $O$ (Figure~\ref{fig:mutag1}). If a graph has at least $4$ such nodes, it is mutagenic (Figure~\ref{fig:mutagtree}).

\label{fig:mutag}

\subsection{REDDIT-BINARY}
\begin{figure}[h!]
    \begin{subfigure}{0.35\textwidth}
        \centering
        \includegraphics[width=\textwidth]{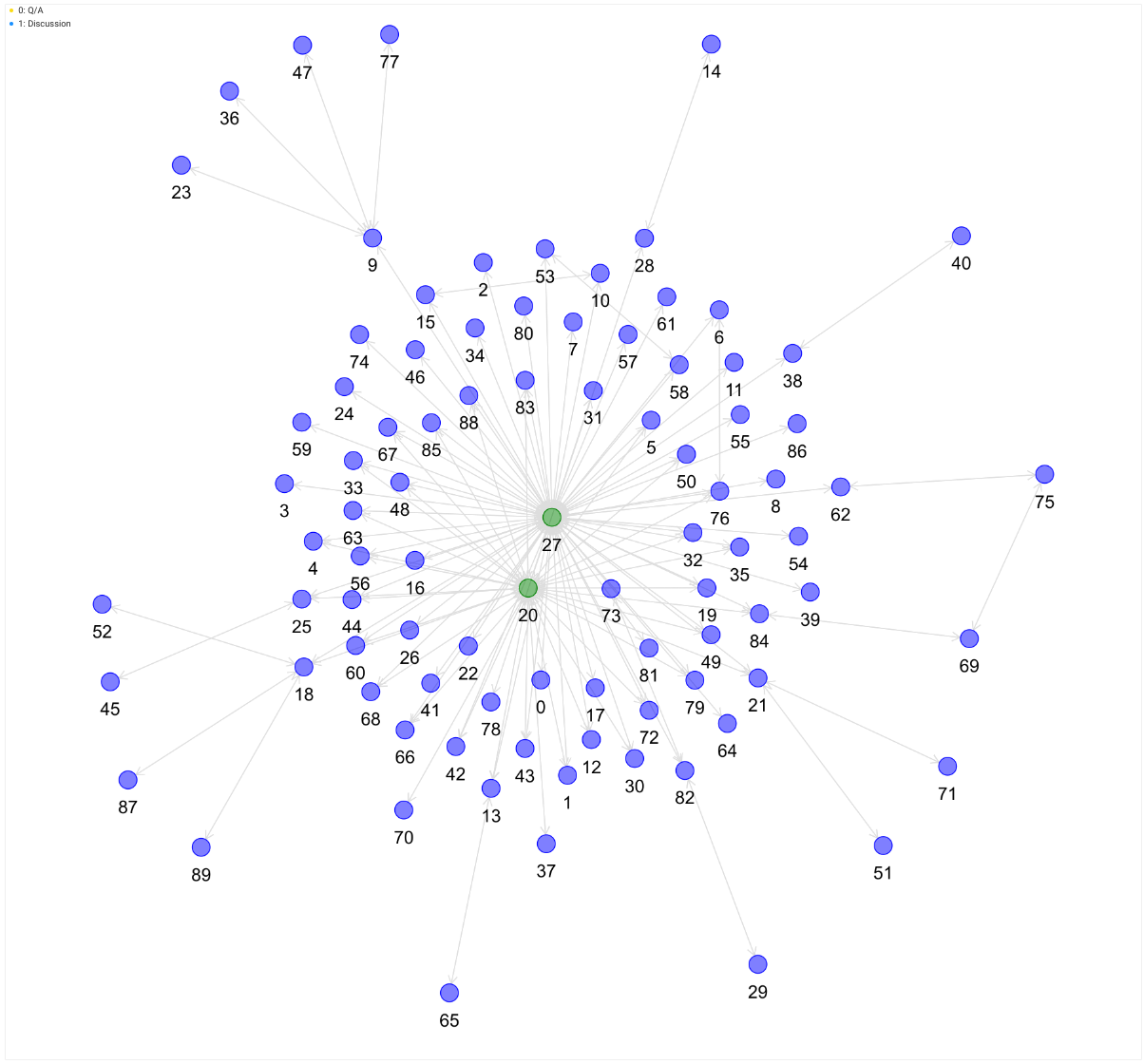}
        \caption{}
        \label{fig:reddit1}
    \end{subfigure}\hfill%
    \begin{subfigure}{0.35\textwidth}
        \centering
        \includegraphics[width=\textwidth]{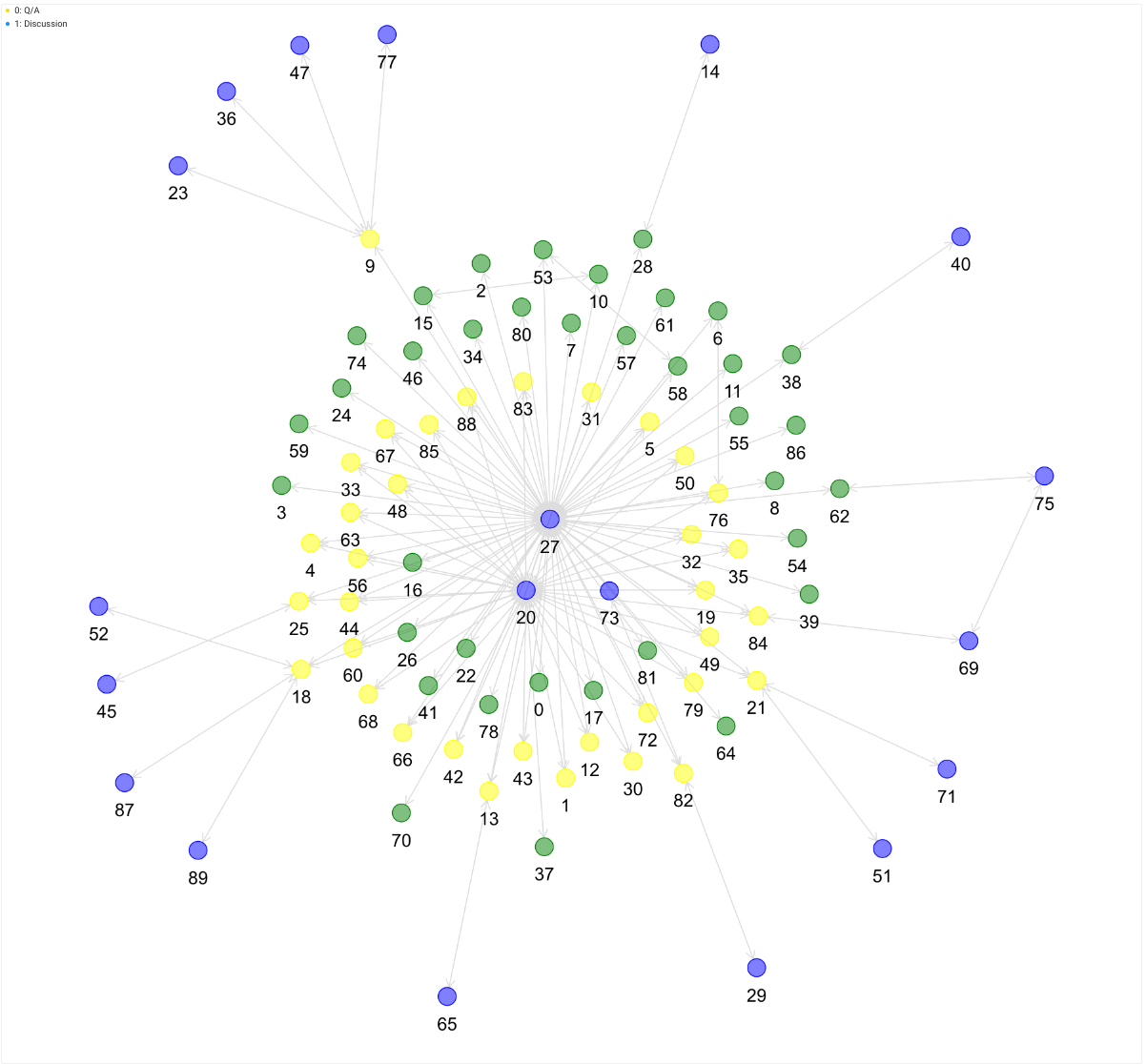}
        \caption{}
        \label{fig:reddit2}
    \end{subfigure}\hfill%
    \begin{subfigure}{0.23\textwidth}
        \centering
        \includegraphics[width=\textwidth]{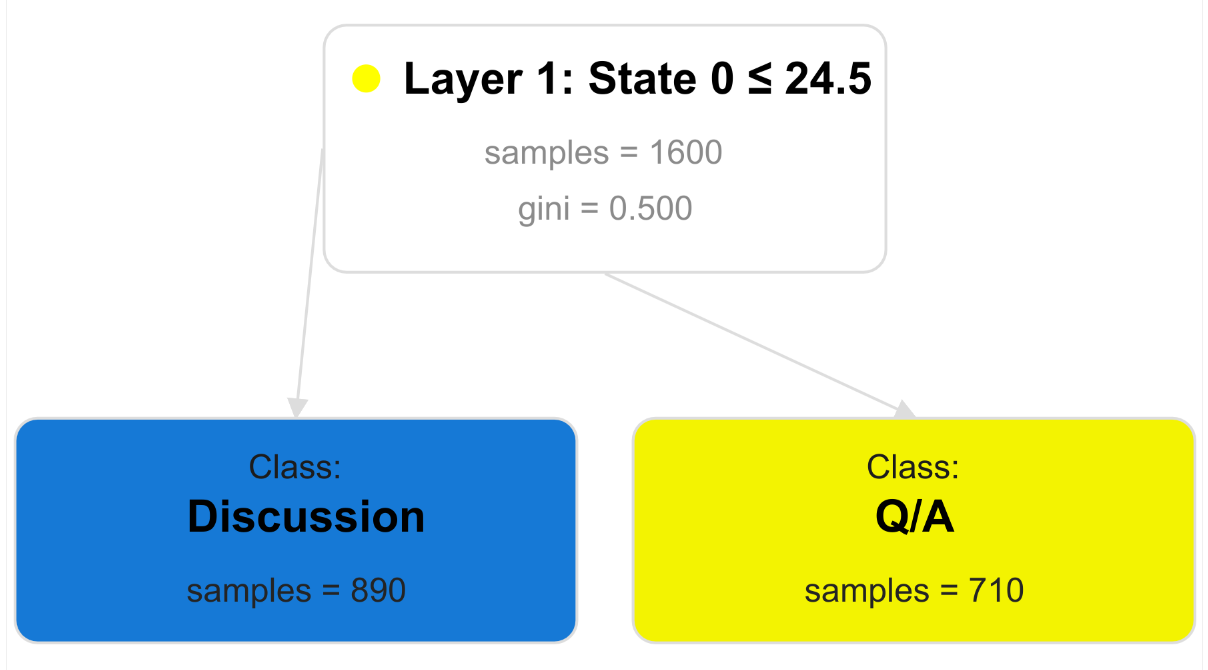}
        \caption{}
        \label{fig:reddittree}
    \end{subfigure}\hfill%
    \caption{DT+GNN solves a ``Q\&A'' instance of the REDDIT-BINARY dataset. (a) and (b) show the learned node states for the first two layers. Numbers denote node IDs. (a) DT+GNN finds the high-degree nodes answering questions. (b) DT+GNN then finds the neighbors of the central nodes that also interact with other nodes (c) shows the decoder decision tree. Since there are $25$ or more such interactive nodes, DT+GNN considers the graph ``Q\&A''.}
\end{figure}

This dataset~\citep{Borgwardt2005ProteinKernels} consists of unattributed graphs that represent a thread in a subreddit. Nodes represent Reddit users; edges are between two users where one commented on another. Depending on the subreddit, the graph has the label ``Discussion'' or ``Q\&A''. \citet{ying2019gnnexplainer} analyzed this dataset before: characteristic of ``Q\&A'' graphs are the high-degree nodes that represent the users answering questions. DT+GNN also finds these users in the first layer (Figure~\ref{fig:reddit1}). Interestingly, DT+GNN bases its prediction on the neighbors of these central nodes. It splits these neighbors into those that interacted with at least $2$ other nodes and those who do not (Figure~\ref{fig:reddit2}). Only if at least $25$ neighbors were interactive, DT+GNN considers the graph a ``Q\&A'' graph (Figure~\ref{fig:reddittree}).
\label{fig:reddit}
\end{document}